%% file: eccv2022submission.tex
\documentclass[runningheads]{llncs}
\usepackage{graphicx}

\usepackage{tikz}
\usepackage{comment}
\usepackage{amsmath,amssymb} 
\usepackage{color}

\usepackage[accsupp]{axessibility}  

\usepackage{kotex}
\usepackage{mathtools}
\usepackage{ulem}
\usepackage{afterpage}
\usepackage{diagbox}
\usepackage{multirow}
\usepackage[colorlinks=true,citecolor={black},linkcolor={black},pagebackref=true,breaklinks=true,bookmarks=false]{hyperref}
\usepackage{breakcites}

\begin{document}
\pagestyle{headings}
\mainmatter
\def\ECCVSubNumber{6325}  

\title{Realistic Blur Synthesis for Learning Image Deblurring} 

\titlerunning{Realistic Blur Synthesis for Learning Image Deblurring}
%
\author{Jaesung Rim \and
Geonung Kim \and
Jungeon Kim \and
Junyong Lee \and \\
Seungyong Lee \and
Sunghyun Cho}
\authorrunning{J. Rim et al.}
\institute{
POSTECH, Pohang, Korea \\
\email{\{jsrim123,k2woong92,jungeonkim,junyonglee,leesy,s.cho\}@postech.ac.kr}
\\
\vspace{1pt}
\tt\small \url{http://cg.postech.ac.kr/research/RSBlur}
\vspace{-10pt}
}

\maketitle

\input{macro.tex}

\input{0.abstract.tex}
\input{1.intro.tex}
\input{2.related_work.tex}
\input{3.method.tex}
\input{4.experiments.tex}
\input{5.conclusion.tex}

{\small
\noindent \textbf{Acknowledgements} This work was supported by Samsung Research Funding \& Incubation Center
of Samsung Electronics under Project Number SRFC-IT1801-05
and
Institute of Information \& communications Technology Planning \& Evaluation (IITP) grants (2019-0-01906, Artificial Intelligence Graduate School Program (POSTECH)) funded by the Korea government (MSIT)
and 
the National Research Foundation of Korea (NRF) grants (2020R1C1C1014863) funded by the Korea government (MSIT).
}

\input{eccv2022_supplementary.tex}

\clearpage
%
%
\bibliographystyle{splncs04}
\bibliography{eccv2022submission.bbl}

\end{document}

%% file: macro.tex
\newcommand{\Eq}[1]  {Eq.\ (\ref{eq:#1})}
\newcommand{\Eqs}[1] {Eqs.\ (\ref{eq:#1})}
\newcommand{\Fig}[1] {Fig.\ \ref{fig:#1}}
\newcommand{\Figs}[1]{Figs.\ \ref{fig:#1}}
\newcommand{\Tbl}[1]  {Table \ref{tbl:#1}}
\newcommand{\Tbls}[1] {Tables \ref{tbl:#1}}
\newcommand{\Sec}[1] {Sec.\ \ref{sec:#1}}
\newcommand{\Secs}[1] {Secs.\ \ref{sec:#1}}
\newcommand{\etal}   {{\textit{et al.}}}
\newcommand{\Etal}   {{\textit{et al.}}}

\newcommand{\setone}[1] {\left\{ #1 \right\}} 
\newcommand{\settwo}[2] {\left\{ #1 \mid #2 \right\}} 

\definecolor{brown}{rgb}{0.65, 0.16, 0.16}
\definecolor{purp}{rgb}{0.65, 0.16, 0.65}
\definecolor{orange}{rgb}{1.0, 0.5, 0.0}
\definecolor{blue}{rgb}{0.0, 0.5, 1.0}
\definecolor{green}{rgb}{0, 0.8, 0}
\definecolor{lgreen}{rgb}{0.6, 0.8, 0}
\definecolor{red}{rgb}{1.0, 0, 0}
\definecolor{darkblue}{rgb}{0, 0.2, 0.6}

\newcommand{\todo}[1]{{\textcolor{blue}{TODO: #1}}}
\newcommand{\son}[1]{{\textcolor{magenta}{lee: #1}}}
\newcommand{\sean}[2]{{\textcolor{green}{sean: #1}\textcolor{magenta}{#2}}}
\newcommand{\sunghyun}[1]{{\textcolor[rgb]{0.6,0.0,0.6}{sunghyun: #1}}}
\newcommand{\rjs}[1]{{\textcolor[rgb]{1,0.0,0.0}{#1}}}
\newcommand{\rjsc}[1]{{\textcolor[rgb]{0.6,0.6,1}{#1}}}
\newcommand{\kyoungkook}[1]{{\textcolor[rgb]{0.6,0.6,1}{kyoungkook: #1}}}
\newcommand{\jy}[1]{{\textcolor[rgb]{0.6,0.6,1}{jy: #1}}}
\newcommand{\change}[1]{{\color{red}#1}}

\renewcommand{\topfraction}{0.95}
\setcounter{bottomnumber}{1}
\renewcommand{\bottomfraction}{0.95}
\setcounter{totalnumber}{3}
\renewcommand{\textfraction}{0.05}
\renewcommand{\floatpagefraction}{0.95}
\setcounter{dbltopnumber}{2}
\renewcommand{\dbltopfraction}{0.95}
\renewcommand{\dblfloatpagefraction}{0.95}

\newcommand{\argmin}{\mathop{\mathrm{argmin}}\limits} 

\renewcommand{\paragraph}[1]{{\vspace{0.2em}\noindent\textbf{#1}~~}}


%% file: 0.abstract.tex
\begin{abstract}
Training learning-based deblurring methods demands a tre-mendous amount of blurred and sharp image pairs. 
Unfortunately, existing synthetic datasets are not realistic enough, and deblurring models trained on them cannot handle real blurred images effectively. 
While real datasets have recently been proposed, they provide limited diversity of scenes and camera settings, and capturing real datasets for diverse settings is still challenging.
To resolve this, this paper analyzes various factors that introduce differences between real and synthetic blurred images.
To this end, we present RSBlur, a novel dataset with real blurred images and the corresponding sharp image sequences to enable a detailed analysis of the difference between real and synthetic blur. 
With the dataset, we reveal the effects of different factors in the blur generation process.
Based on the analysis, we also present a novel blur synthesis pipeline to synthesize more realistic blur. 
We show that our synthesis pipeline can improve the deblurring performance on real blurred images.

\keywords{Realistic Blur Synthesis, Dataset and Analysis, Deblurring}
\end{abstract}

%% file: 1.intro.tex
\section{Introduction}
\label{sec:intro}

Motion blur is caused by camera shake or object motion during exposure, especially in a low-light environment that requires long exposure time.
Image deblurring is the task of enhancing image quality by removing blur.
For the past several years, numerous learning-based deblurring methods have been introduced and significantly improved the performance~\cite{Nah_2017_CVPR,Tao-CVPR18,DeblurGAN,DeblurGAN-v2,Zhang-CVPR19,Zamir_2021_CVPR,Cho_2021_ICCV,Wang_2022_CVPR,Zamir_2022_CVPR}.

Training learning-based deblurring methods demands a significant amount of blurred and sharp image pairs.
Since it is hard to obtain real-world blurred and sharp image pairs, a number of synthetically generated datasets have been proposed, whose blurred images are generated by blending sharp video frames captured by high-speed cameras \cite{Nah_2017_CVPR,Nah_NTIRE19,Su-CVPR17,Zhou_2019_CVPR,Shen_ICCV19,Deng_2021_ICCV,Li_2021_CVPR}.
Unfortunately, such synthetic images are not realistic enough, so deblurring methods trained on them often fail to deblur real blurred images~\cite{jsrim-ECCV2020}.

To overcome such a limitation, Rim~\etal~\cite{jsrim-ECCV2020} and Zhong~\etal~\cite{Zhong_2020_ECCV,Zhong_2021_arxiv} recently presented the RealBlur and BSD datasets, respectively. These datasets consist of real blurred and sharp ground truth images captured using specially designed dual-camera systems.
Nevertheless, coverage of such real datasets are still limited.
Specifically, both RealBlur and BSD datasets are captured using \emph{a single camera model}, Sony A7R3, and a machine vision camera, respectively~\cite{jsrim-ECCV2020,Zhong_2020_ECCV,Zhong_2021_arxiv}.
As a result, deblurring models trained on each of them show significantly low performance on the other dataset, as shown in \Sec{experiments}.
Moreover, it is not easy to expand the coverage of real datasets as collecting such datasets require specially designed cameras and a tremendous amount of time.

In this paper, we explore ways to synthesize more realistic blurred images for training deblurring models
so that we can improve deblurring quality on real blurred images without the burden of collecting a broad range of real datasets.
To this end, we first present \emph{RSBlur}, a novel dataset of real and synthetic blurred images.
Then, using the dataset, we analyze the difference between the generation process of real and synthetic blurred images and present a realistic blur synthesis method based on the analysis.

Precise analysis of the difference between real and synthetic blurred images requires pairs of synthetic and real blurred images to facilitate isolating factors that cause the difference.
However, there exist no datasets that provide both synthetic and real blurred images of the same scenes so far.
Thus, to facilitate the analysis, the \emph{RSBlur} dataset provides pairs of a real blurred image and a sequence of sharp images captured by a specially-designed high-speed dual-camera system.
With the dataset, we can produce a synthetic blurred image by averaging a sequence of sharp images and compare it with its corresponding real blurred image.
This allows us to analyze the difference between real and synthetic blurred images focusing on their generation processes.
In particular, we investigate several factors that may degrade deblurring performance of synthetic datasets on real blurred images, such as noise, saturated pixels, and camera ISP.
Based on the analysis, we present a method to synthesize more realistic blurred images.
Our experiments show that our method can synthesize more realistic blurred images, and our synthesized training set can greatly improve the deblurring performance on real blurred images compared to existing synthetic datasets.

Our contributions are summarized as follows:
\begin{itemize}
\item We propose \emph{RSBlur}, the first dataset that provides pairs of a real blurred image and a sequence of sharp images, which enables accurate analysis of the difference between real and synthetic blur.
\vspace{+0.4em}
\item We provide a thorough analysis of the difference between the generation processes of real and synthetic blurred images.
\vspace{+0.4em}
\item We present a novel synthesis method to synthesize realistic blurred images for learning image deblurring. 
While collecting large-scale real datasets for different cameras is challenging,
our method offers a convenient alternative.
\end{itemize}

%% file: 2.related_work.tex
\vspace{-5pt}
\section{Related Work}

\paragraph{Deblurring Methods}
Traditional deblurring methods rely on restrictive blur models,
thus they often fail to deblur real-world images \cite{Shan-SIGGRAPH08,Cho-SIGGRAPHAsia09,Xu-ECCV10,Pan-CVPR16,Sun-ICCP13,Cho-ICCV17,Levin-CVPR09,Levin-CVPR11}.
To overcome such limitations, learning-based approaches that restore a sharp image from a blurred image by learning from a large dataset have recently been proposed~\cite{Nah_2017_CVPR,Tao-CVPR18,DeblurGAN,DeblurGAN-v2,Zhang-CVPR19,Zamir_2021_CVPR,Cho_2021_ICCV,Wang_2022_CVPR,Zamir_2022_CVPR}.
However, they require a large amount of training data.

\paragraph{Deblurring Datasets}
For evaluation of deblurring methods, Levin \etal~\cite{Levin-CVPR09} and K\"{o}hler \etal~\cite{Kohler-ECCV12} collected real blurred images by capturing images on the wall while shaking the cameras. Sun \etal~\cite{Sun-ICCP13} generate 640 synthetic blurred images by convolving 80 sharp images with eight blur kernels. Lai \etal~\cite{Lai-CVPR16} generate spatially varying blurred images from 6D camera trajectories and construct a dataset including 100 real blurred images. However, due to the small number of images, these datasets cannot be used for learning-based methods.

Several synthetic datasets using high-speed videos have been proposed for training learning-based methods. They capture high-speed videos and generate synthetic blurred images by averaging sharp frames.
GoPro \cite{Nah_2017_CVPR} is the most widely used dataset for learning-based deblurring methods.
REDS \cite{Nah_NTIRE19} and DVD \cite{Su-CVPR17} provide synthetically blurred videos for learning video deblurring.
Stereo Blur \cite{Zhou_2019_CVPR} consists of stereo blurred videos generated by averaging high-speed stereo video frames.
HIDE \cite{Shen_ICCV19} provides synthetic blurred images with bounding box labels of humans.
To expand deblurring into high-resolution images, 4KRD~\cite{Deng_2021_ICCV} is presented, which consists of synthetically blurred UHD video frames.
All the datasets discussed above, except for GoPro, use frame interpolation before averaging sharp images to synthesize more realistic blur \cite{Nah_NTIRE19,Su-CVPR17,Zhou_2019_CVPR,Shen_ICCV19,Deng_2021_ICCV}.
HFR-DVD \cite{Li_2021_CVPR} uses high-speed video frames captured at 1000 FPS to synthesize blur without frame interpolation.
However, all the aforementioned datasets are not realistic enough, thus deblurring networks trained with them often fail to deblur real-world blurred images.

Recently, real-world blur datasets \cite{jsrim-ECCV2020,Zhong_2020_ECCV,Zhong_2021_arxiv} have been proposed.
They simultaneously capture a real blurred image and its corresponding sharp image using a dual-camera system. Rim \etal~\cite{jsrim-ECCV2020} collected a real-world blur dataset in low-light environments.
Zhong \etal~\cite{Zhong_2020_ECCV,Zhong_2021_arxiv} proposed the BSD dataset containing pairs of real blurred and sharp videos.
However, their performances degrade on other real images captured in different settings due to their limited coverage. 

\paragraph{Synthesis of Realistic Degraded Images}
In the denoising field, synthesizing more realistic noise for learning real-world denoising has been actively studied~\cite{Abdelhamed_2019_ICCV,Chang_2020_ECCV,Jang_2021_ICCV,Wei_2020_CVPR,Guo_2019_CVPR,Brooks_2019_CVPR_denoising}.
Abdelhamed \etal~\cite{Abdelhamed_2019_ICCV}, Chang \etal~\cite{Chang_2020_ECCV}, and Jang \etal~\cite{Jang_2021_ICCV} use generative models to learn a mapping from a latent distribution to a real noise distribution.
Zhang \etal~\cite{Zhang_2021_ICCV} and Wei \etal~\cite{Wei_2020_CVPR} propose realistic noise generation methods based on the physical properties of digital sensors.
Guo \etal~\cite{Guo_2019_CVPR} and Brooks \etal~\cite{Brooks_2019_CVPR_denoising} generate realistic noise by unprocessing arbitrary clean sRGB images, adding Poisson noise, and processing them back to produce noisy sRGB images.
These methods show that more realistically synthesized noise datasets greatly improve the denoising performance of real-world noisy images.

\paragraph{Synthesis of Blurred Images}
A few methods have been proposed to synthesize blur without using high-speed videos~\cite{Brooks_2019_CVPR,Zhang_2020_CVPR}.
Brooks \etal~\cite{Brooks_2019_CVPR} generate a blurred image from two sharp images using a line prediction layer, which estimates spatially-varying linear blur kernels.
However, linear blur kernels cannot express a wide variety of real-world blur.
Zhang \etal~\cite{Zhang_2020_CVPR} use real-world blurred images without their ground-truth sharp images to train a GAN-based model to generate a blurred image from a single sharp image.
However, their results are not realistic enough as their generative model cannot accurately reflect the physical properties of real-world blur.

%% file: 3.method.tex
\begin{figure}[!t]
\begin{center}
\includegraphics [width=0.9\linewidth] {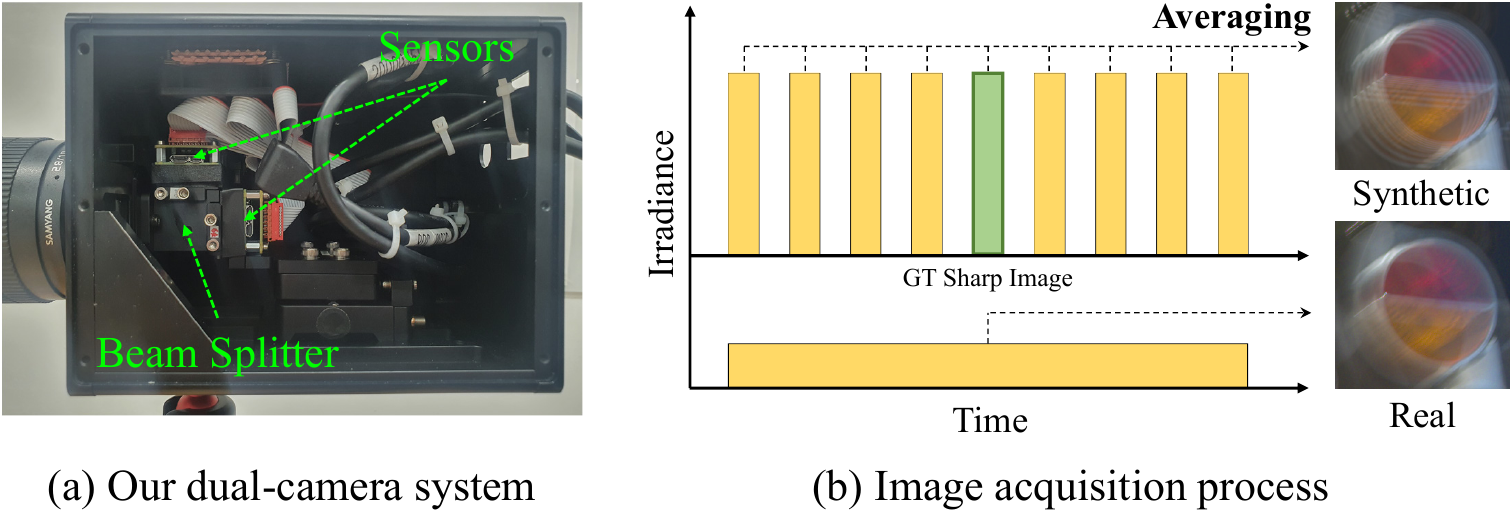}
\end{center}
\vspace{-0.65cm}
\caption{The dual-camera system and the acquisition process for simultaneously capturing a real blurred image and sharp images.}
\label{fig:our_system}
\vspace{-0.45cm}
\end{figure}

\vspace{-5pt}
\section{RSBlur Dataset}
\label{sec:rsblur}

Our proposed RSBlur dataset provides real blurred images of various outdoor scenes, each of which is paired with a sequence of nine sharp images to enable the analysis of the difference between real and synthetic blur.
The dataset includes a total of 13,358 real blurred images of 697 scenes.
For our analysis, we split the dataset into training, validation, and test sets with 8,878, 1,120, and 3,360 blurred images of 465, 58, and 174 scenes, respectively.
Below, we explain the acquisition process and other details of the RSBlur dataset.

To collect the dataset, we built a dual camera system (\Fig{our_system}(a)) as done in \cite{jsrim-ECCV2020,Zhong_2020_ECCV,Zhong_2021_arxiv,Zhong_2021_CVPR}.
The system consists of one lens, one beam splitter, and two camera modules with imaging sensors so that the camera modules can capture the same scene while sharing one lens.
The shutters of the camera modules are carefully synchronized in order that the modules can capture images simultaneously (\Fig{our_system}(b)).
Specifically, one camera module captures a blurred image with a long exposure time.
During the exposure time of a blurred image, the other module captures nine sharp images consecutively with a short exposure time.
The shutter for the first sharp image opens when the shutter for the blurred image opens,
and the shutter for the last sharp image closes when the shutter for the blurred image closes.
The exposure time of the fifth sharp image matches with the center of the exposure time of the blurred image so that the fifth sharp image can be used as a ground-truth sharp image for the blurred one.

The blurred images are captured with a 5\% neutral density filter installed in front of a camera module to secure a long exposure time as done in \cite{Zhong_2020_ECCV,Zhong_2021_arxiv}.
The exposure times for the blurred and sharp images are 0.1 and 0.005 seconds, respectively.
We capture images holding our system in hand so that blurred images can be produced by hand shakes.
The captured images are geometrically and photometrically aligned to remove misalignment between the camera modules as done in \cite{jsrim-ECCV2020}. 
We capture all images in the camera RAW format, and convert them into the nonlinear sRGB space using a simple image signal processing (ISP) pipeline similar to \cite{Abdelhamed_2018_CVPR} consisting of four steps: 1) white balance, 2) demosaicing, 3) color correction, and 
4) conversion to the sRGB space using a gamma correction of sRGB space as a camera response function (CRF).
More details on our dual-camera system and ISP are in the supplement.

\vspace{-5pt}
\section{Real \textbf{\textit{vs}} Synthetic Blur}
\label{sec:real_vs_syn}

Using the RSBlur dataset, we analyze the difference between the generation process of real and synthetic blur.
Specifically, we first compare the overall generation process of real and synthetic blur, and discover factors that can introduce the dominant difference between them.
Then, we analyze each factor one by one and discuss how to address them by building our blur synthesis pipeline.

In the case of real blur, camera sensors accumulate incoming light during the exposure time to capture an image. During this process, blur and photon shot noise are introduced due to camera and object motion, and due to the fluctuation of photons, respectively.
The limited dynamic range of sensors introduces saturated pixels.
The captured light is converted to analog electrical signals and then to digital signals. During this conversion, additional noise such as dark current noise and quantization noise is added.
The image is then processed by a camera ISP, which performs white balance, demosaicing, color space conversion, and other nonlinear operation that distort the blur pattern and noise distribution.

During this process, an image is converted through multiple color spaces.
Before the camera ISP, an image is in the camera RAW space, which is device-dependent.
The image is then converted to the linear sRGB space, and then to the nonlinear sRGB space.
In the rest of the paper, we refer to the linear sRGB space as the linear space, and the nonlinear sRGB space as the sRGB space.

On the other hand, the blurred image generation processes of the widely used datasets, e.g., GoPro~\cite{Nah_2017_CVPR}, DVD~\cite{Su-CVPR17}, and REDS~\cite{Nah_NTIRE19}, are much simpler.
They use sharp images in the sRGB space consecutively captured by a high-speed camera.
The sharp images are optionally interpolated to increase the frame rate~\cite{Su-CVPR17,Nah_NTIRE19}.
Then, they are converted to the linear space, and averaged together to produce a blurred image.
The blurred image is converted to the sRGB space.
For conversion between the linear to sRGB spaces, GoPro uses a gamma curve with $\gamma=2.2$ while REDS uses a CRF estimated from a GOPRO6 camera. 

Between the two processes described above, the main factors that cause the gap between synthetic and real blur include 1) discontinuous blur trajectories in synthetic blur, 2) saturated pixels, 3) noise, and 4) the camera ISP.
In this paper, we analyze the effect of these factors one by one.
Below, we discuss these factors in more detail.

\begin{figure}[!t]
\centering
\includegraphics[width=0.75\linewidth]{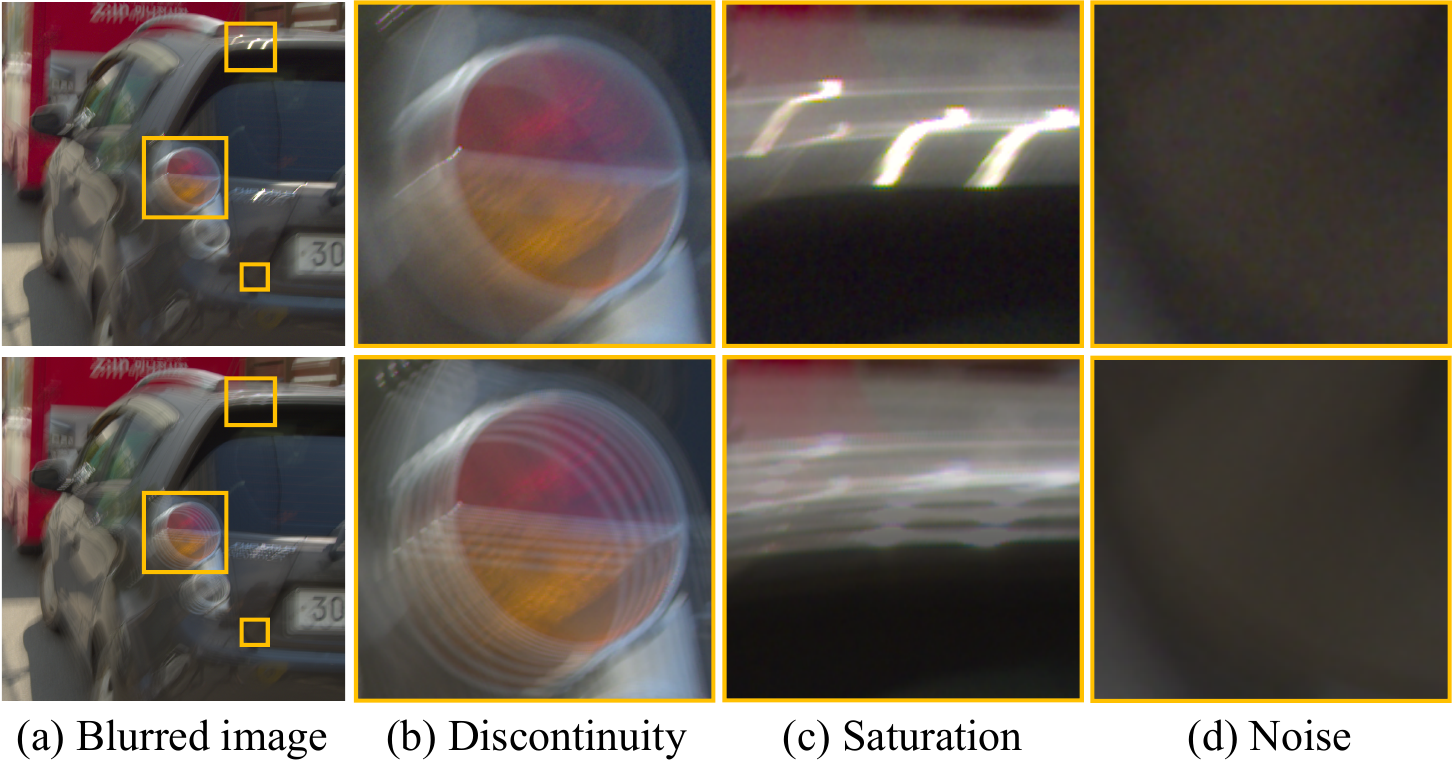}
\vspace{-0.25cm}
\caption{The top row shows real blurred images and the bottom row shows the corresponding synthetic blurred images. Best viewed in zoom in.}
\vspace{-0.45cm}
\label{fig:real_vs_syn}
\end{figure}

\paragraph{Discontinuous Blur Trajectories}
The blur generation process of the GoPro dataset~\cite{Nah_2017_CVPR}, which is the most popular dataset, captures sharp video frames at a high frame rate and averages them to synthesize blur.
However, temporal gaps between the exposure of consecutive frames cause 
unnatural discontinuous blur (\Fig{real_vs_syn}(b)).
While DVD~\cite{Su-CVPR17} and REDS~\cite{Nah_NTIRE19} use frame interpolation to fill such gaps, the effects of discontinuous blur and frame interpolation on the deblurring performance have not been analyzed yet.

\paragraph{Saturated Pixels}
While real-world blurred images may have saturated pixels (\Fig{real_vs_syn}(c)) due to the limited dynamic range,
previous synthetic datasets do not have such saturated pixels as they simply average sharp images.
As saturated pixels in real blurred images form distinctive blur patterns from other pixels, it is essential to reflect them to achieve high-quality deblurring results \cite{Cho-ICCV11}.

\paragraph{Noise}
Noise is inevitable in real-world images including blurred images, especially captured by a low-end camera at night (\Fig{real_vs_syn}(d)).
Even for high-end sensors, noise cannot be avoided due to the statistical
property of photons and the circuit readout process.
In the denoising field, it has been proven important to model the realistic noise for high-quality denoising of real-world images~\cite{Zhang_2021_ICCV,Wei_2020_CVPR,Abdelhamed_2019_ICCV,Chang_2020_ECCV,Jang_2021_ICCV,Brooks_2019_CVPR_denoising}.
On the other hand, noise is ignored by the blur generation processes of the previous synthetic datasets~\cite{Nah_2017_CVPR,Nah_NTIRE19,Su-CVPR17,Zhou_2019_CVPR,Shen_ICCV19,Deng_2021_ICCV,Li_2021_CVPR}, and its effect on deblurring has not been investigated.
Our experiments in \Sec{experiments} show that accurate modeling of noise is essential 
even for the RealBlur dataset, which consists of images mostly captured from a high-end camera with the lowest ISO.

\paragraph{Camera ISP}
ISPs perform various operations, including white balancing, color correction, demosaicing, and nonlinear mapping using CRFs, which affect the noise distribution and introduce distortions~\cite{Brooks_2019_CVPR_denoising,Yue_2021_arxiv}. However, they are ignored by the previous synthetic datasets~\cite{Nah_2017_CVPR,Nah_NTIRE19,Su-CVPR17,Zhou_2019_CVPR,Shen_ICCV19,Deng_2021_ICCV,Li_2021_CVPR}. 

\vspace{-5pt}
\section{Realistic Blur Synthesis}
\label{sec:blur_synthesis}

\begin{figure}[!t]
\centering
\includegraphics[width=0.95\linewidth]{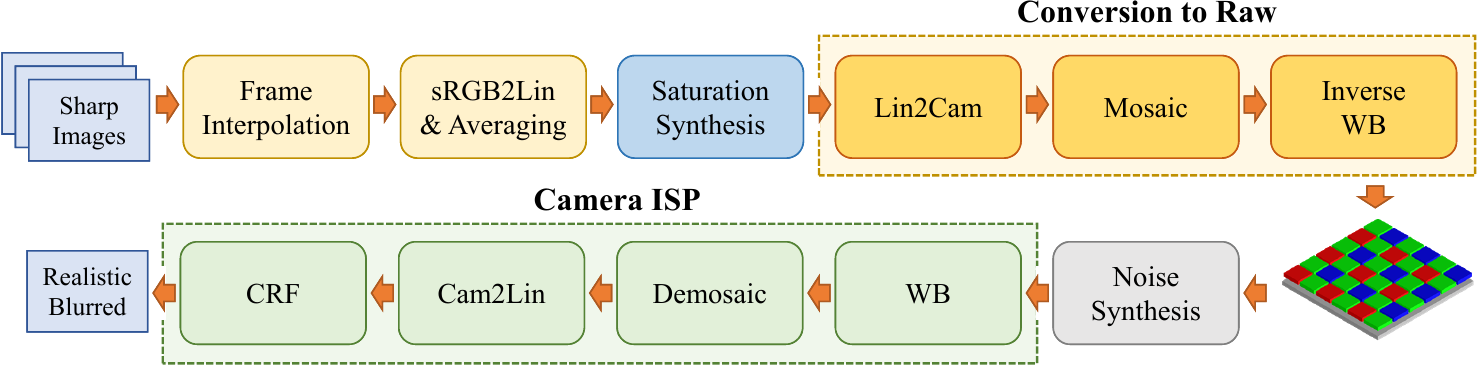}
\vspace{-0.15cm}
\caption{Overview of our realistic blur synthesis pipeline. Lin2Cam: Inverse color correction, i.e., color space conversion from the linear space to the camera RAW space. WB: White balance. Cam2Lin: Color correction.}
\vspace{-0.45cm}
\label{fig:generation_process}
\end{figure}

To synthesize more realistic blur while addressing the factors discussed earlier, we propose a novel blur synthesis pipeline.
The proposed pipeline will also serve as a basis for the experiments in \Sec{experiments} that study effect of each factor that degrades the quality of synthetic blur.
\Fig{generation_process} shows an overview of our blur synthesis pipeline.
Our pipeline takes sharp video frames captured by a high-speed camera as done in \cite{Nah_2017_CVPR,Nah_NTIRE19,Su-CVPR17}, and produces a synthetic blurred image.
Both input and output of our pipeline are in the sRGB space.
Below, we explain each step in more detail.

\paragraph{Frame Interpolation}
To resolve the discontinuity of blur trajectory, our pipeline adopts frame interpolation as done in \cite{Su-CVPR17,Nah_NTIRE19}.
We increase nine sharp images to 65 images using ABME~\cite{Park_2021_ICCV}, a state-of-the-art frame interpolation method.
In this step, we perform frame interpolation in the sRGB space to use an off-the-shelf frame interpolation method without modification or fine-tuning.

\begin{figure}[!t]
\centering
\includegraphics[width=1\linewidth]{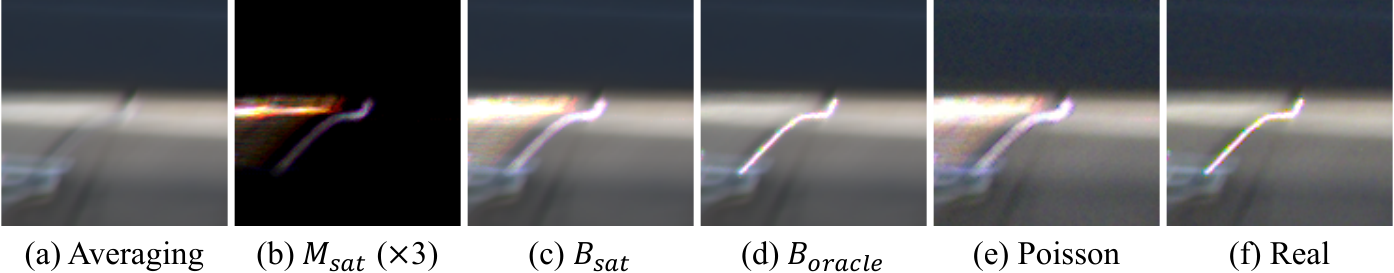}
\vspace{-0.65cm}
\caption{Generated images from our synthesis pipeline. (a) Averaging image of interpolated frames. (b) $M_{sat}$ scaled by three times. (c)-(d) Examples of saturated images. (e)-(f) Synthetic noisy image and real image. The images except for (b) are converted into the sRGB space for visualization.}
\vspace{-0.45cm}
\label{fig:generating_results}
\end{figure}

\paragraph{sRGB2Lin \& Averaging}
To synthesize blur using the interpolated frames,
we convert the images into the linear space, and average them to precisely mimic the real blur generation process.
While the actual accumulation of incoming light happens in the camera RAW space, averaging in the camera RAW space and in the linear space are equivalent to each other as the two spaces can be converted using a linear transformation.
\Fig{generating_results}(a) shows an example of the averaging of interpolated frames.

\paragraph{Saturation Synthesis}
In this step, we synthesize saturated pixels.
To this end, we propose a simple approach.
For a given synthetic blurred image $B_{syn}$ from the previous step, our approach first calculates a mask $M_i$ of the saturated pixels in the $i$-th sharp source image $S_i$ of $B_{syn}$ as follows: 
\begin{equation} \label{eq1}
  M_i(x,y,c) = \begin{cases*} 1,  & if $ S_i(x,y,c) = 1 $ \\
                           0,  & otherwise,
             \end{cases*}
\end{equation}
where $(x,y)$ is a pixel position, and $c\in\{R,G,B\}$ is a channel index.
$S_i$ has a normalized intensity range $[0,1]$.
Then, we compute a mask $M_{sat}$ of potential saturated pixels in $B_{syn}$ by averaging $M_i$'s.
\Fig{generating_results}(b) shows an example of $M_{sat}$.
Using $M_{sat}$, we generate a blurred image $B_{sat}$ with saturated pixels as:
\begin{equation}
    B_{sat} = \textrm{clip}(B_{syn} + \alpha M_{sat})
\end{equation}
where $\textrm{clip}(\cdot)$ is a clipping function that clips input values into $[0, 1]$, and $\alpha$ is a scaling factor randomly sampled from a uniform distribution $\mathcal{U}(0.25, 1.75)$.

For the sake of analysis, we also generate blurred images with oracle saturated pixels. An oracle image $B_{oracle}$ is generated as:
\begin{equation} \label{B_oracle}
 B_{oracle}(x,y,c) = \begin{cases*} B_{real}(x,y,c),  & if $ M_{sat}(x,y,c) > 0 $ \\
                         B_{syn}(x,y,c),  & otherwise.
            \end{cases*}
\end{equation}
Our approach is simple and heuristic, and cannot reproduce the saturated pixels in real images due to missing information in sharp images.
Specifically, while we resort to a randomly-sampled uniform scaling factor $\alpha$, for accurate reconstruction of saturated pixels, we need pixel-wise scaling factors, which are impossible to estimate.
\Fig{generating_results}(c) and (d) show examples of $B_{sat}$ and $B_{oracle}$ where the image in (c) looks different from the one in (d).
Nevertheless, our experiments in \Sec{experiments} show that our approach still noticeably improves the deblurring performance on real blurred images.

\paragraph{Conversion to RAW}
In the next step, we convert the blurred image from the previous step, which is in the linear space, into the camera RAW space to reflect the distortion introduced by the camera ISP.
In this step, we apply the inverse of each step of our ISP except for the CRF step in the reverse order.
Specifically, we apply the inverse color correction transformation, mosaicing, and inverse white balance sequentially.
As the color correction and white balance operations are invertible linear operations, they can be easily inverted.
More details are provided in the supplement.

\paragraph{Noise Synthesis}
After the conversion to the camera RAW space, we add noise to the image.
Motivated by \cite{Wei_2020_CVPR,Zhang_2021_ICCV}, we model noise in the camera RAW space as a mixture of Gaussian and Poisson noise as:
\begin{equation}
    B_{noisy} = \beta_1 ( I + N_{shot} ) + N_{read}
\end{equation}
where $B_{noisy}$ is a noisy image, and $I$ is the number of incident photons.
$\beta_1$ is the overall system gain determined by digital and analog gains.
$N_{shot}$ and $N_{read}$ are photon shot and read noise, respectively.
We model $(I+N_{shot})$ as a Poisson distribution, and $N_{read}$ as a Gaussian distribution with standard deviation $\beta_2$.
Mathematically, $(I+N_{shot})$ and $N_{read}$ are modeled as:
\begin{eqnarray}
    (I+N_{shot}) &\sim&  \mathcal{P}\left (\frac{B_{raw}}{\beta_1} \right) \beta_1,~~~\textrm{and} \label{eq:shot_noise}\\
    N_{read} &\sim& \mathcal{N}(0, \beta_2) \label{eq:read_noise}
\end{eqnarray}
where $\mathcal{P}$ and $\mathcal{N}$ denote Poisson and Gaussian distributions, respectively.
$B_{raw}$ is a blurred image in the camera RAW space from the previous step.

To reflect the noise distribution in the blurred images in the RSBlur dataset, we estimate the parameters $\beta_1$ and $\beta_2$ of our camera system as done in \cite{Zhang_2021_ICCV},
where $\beta_1$ and $\beta_2$ are estimated using flat-field and dark-frame images, respectively. Refer to \cite{Zhang_2021_ICCV} for more details.
The estimated values of $\beta_1$ and $\beta_2$ are 0.0001 and 0.0009, respectively.
To cover a wider range of noise in our synthetic blurred images,
we sample random parameter values $\beta_1'$ and $\beta_2'$ from 
$\mathcal{U}(0.5\beta_1, 1.5\beta_1)$ and $\mathcal{U}(0.5\beta_2, 1.5\beta_2)$, respectively.
Then, using \Eq{shot_noise} and \Eq{read_noise} with $\beta_1'$ and $\beta_2'$, we generate a noisy blurred image in the camera RAW space.

For the analysis in \Sec{experiments}, we also consider Gaussian noise, which is the most widely used noise model.
We obtain a noisy image with Gaussian noise as:
\begin{equation}
B_{noisy} = B + N_{gauss}    
\end{equation}
where $B$ is an input blurred image and $N_{gauss}$ is Gaussian noise sampled from $\mathcal{N}(0, \sigma)$, and $\sigma$ is the standard deviation.
As we include Gaussian noise in our analysis to represent the conventional noise synthesis, we skip the ISP-related steps (conversion to RAW, and applying camera ISP), but directly add noise to a blurred image in the sRGB space, i.e., we apply gamma correction to $B_{sat}$ from the previous step, and add Gaussian noise to produce the final results.
In our experiments, we randomly sample standard deviations of Gaussian noise from $\mathcal{U}(0.5\sigma',1.5\sigma')$ where $\sigma'=0.0112$ is estimated using a color chart image.

\paragraph{Applying Camera ISP}
Finally, after adding noise, we apply the camera ISP to the noisy image to obtain a blurred image in the sRGB space.
We apply the same ISP described in \Sec{rsblur}, which consists of white balance, demosaicing, color correction, and CRF steps.
\Fig{generating_results}(e) shows our synthesis result with Poisson noise and ISP distortions. As the example shows, our synthesis pipeline can synthesize a realistic-looking blurred image.

%% file: 4.experiments.tex
\vspace{-5pt}
\section{Experiments}
\label{sec:experiments}

In this section, we evaluate the performance of our blur synthesis pipeline, and the effect of its components on the RSBlur and other datasets.
To this end, we synthesize blurred images using variants of our pipeline, and train a learning-based deblurring method using them. We then evaluate its performance on real blur datasets.
In our analysis, we use SRN-DeblurNet \cite{Tao-CVPR18} as it is a strong baseline~\cite{jsrim-ECCV2020}, and requires a relatively short training time.
We train the model for 262,000 iterations, which is half the iterations suggested in \cite{Tao-CVPR18}, with additional augmentations including random horizontal and vertical flip, and random rotation, which we found improve the performance. We also provide additional analysis results using another state-of-the-art deblurring method, MIMO-UNet~\cite{Cho_2021_ICCV}, in the supplement.

\subsection{Analysis using the RSBlur Dataset}
\label{sec:analysis_using_RSBlur}

We first evaluate the performance of the blur synthesis pipeline, and analyze the effect of our pipeline using the RSBlur dataset.
\Tbl{analysis_result} compares different variants of our blur synthesis pipeline.
The method 1 uses real blurred images for training SRN-DeblurNet model~\cite{Tao-CVPR18}, while the others use synthetic images for training.
To study the effect of saturated pixels, we divide the RSBlur test set into two sets, one of which consists of images with saturated pixels, and the other does not, based on whether a blurred image has more than 1,000 non-zero pixels in $M_{sat}$ computed from its corresponding sharp image sequence.
The numbers of images in the sets with and without saturated pixels are 1,626 and 1,734, respectively.
Below, we analyze the effects of different methods and components based on \Tbl{analysis_result}.
As the table includes a large number of combinations of different components, we include the indices of methods that each analysis compares in the title of each paragraph.

\afterpage{
\setlength{\tabcolsep}{3pt}
\begin{table}[t]
\centering
\caption{Performance comparison among different blur synthesis methods on the RSBlur test set. Interp.: Frame interpolation. Sat.: Saturation synthesis. sRGB: Gamma correction of sRGB space. G: Gaussian noise. G+P: Gaussian and Poisson noise.}
\label{tbl:analysis_result}
\vspace{-0.3cm}
\scalebox{0.9}{
\begin{tabular}{|ccccccc|c|c|c|}
\hline
\multicolumn{7}{|c|}{Blur Synthesis Methods}                                                      & \multicolumn{3}{c|}{PSNR / SSIM}                 \\ \hline
No. & Real                      & CRF    & Interp.             & Sat. & Noise & ISP     & All & Saturated      & No Saturated            \\ \hline
1 & \checkmark &        &                           &            &       &        & 32.53 / 0.8398 & 31.20 / 0.8313 & 33.78 / 0.8478  \\
2 &                          & Linear &                           &            &       &         & 30.12 / 0.7727 & 28.67 / 0.7657 & 31.47 / 0.7793 \\
3 &                          & sRGB    &                           &            &       &         & 30.90 / 0.7805 & 29.60 / 0.7745 & 32.13 / 0.7861 \\
4 &                           & sRGB    &            &            & G     &         & 31.69 / 0.8258 & 30.18 / 0.8174 & 33.11 / 0.8336 \\
5 &                           & sRGB    & \checkmark &            &       &         & 30.20 / 0.7468 & 29.06 / 0.7423 & 31.27 / 0.7511 \\
6 &                           & sRGB    & \checkmark &            & G     &         & 31.77 / 0.8275 & 30.28 / 0.8194 & 33.17 / 0.8352 \\
7 &                          & sRGB    & \checkmark & Oracle     & G     &         & 31.89 / 0.8267 & 30.58 / 0.8191 & 33.12 / 0.8338 \\
8 &                          & sRGB    & \checkmark & Ours    & G     &         & 31.83 / 0.8265 & 30.47 / 0.8187 & 33.12 / 0.8339 \\
9 &                          & sRGB    & \checkmark & Oracle     & G+P   & \checkmark   & 32.06 / 0.8315 & 30.79 / 0.8243 & 33.25 / 0.8384 \\
10 &                          & sRGB    & \checkmark & Ours   & G+P   & \checkmark   & 32.06 / 0.8322 & 30.74 / 0.8248 & 33.30 / 0.8391 \\
\hline
\end{tabular}
}
\vspace{-0.3cm}
\end{table}

\begin{figure}[t]
\centering
\includegraphics[width=1\linewidth]{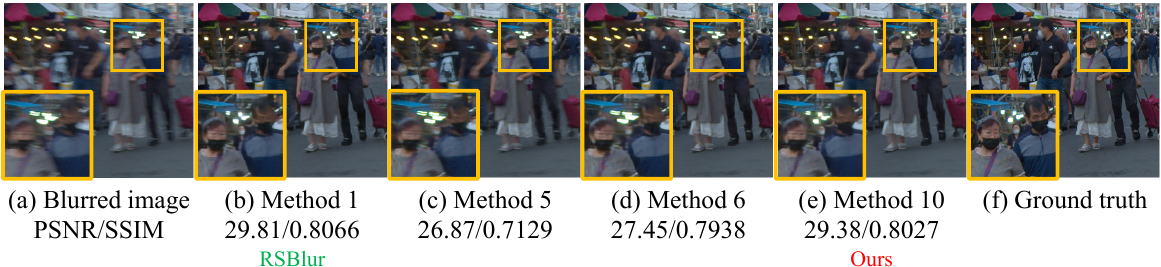}
\vspace{-0.75cm}
\caption{Qualitative comparison of deblurring results on the RSBlur test set produced by models trained with different synthesis methods.
(b)-(e) Methods 1, 5, 6 and 10 in \Tbl{analysis_result}.
Best viewed in zoom in.}
\vspace{-0.3cm}
\label{fig:qualitative_results}
\end{figure}

}

\paragraph{Na\"{i}ve Averaging (2 \& 3)}
We first evaluate the performance of the na\"{i}ve averaging approach, which is used in the GoPro dataset~\cite{Nah_2017_CVPR}.
The GoPro dataset provides two sub-datasets: one of which applies gamma-decoding and encoding before and after averaging, and the other performs averaging without gamma-decoding and encoding.
Thus, in this analysis, we also include two versions of na\"{i}ve averaging.
The method 2 in \Tbl{analysis_result} is the most na\"{i}ve approach, which uses na\"{i}ve averaging and ignores CRFs.
The method 3 also uses na\"{i}ve averaging, but it uses a gamma correction of sRGB space as a CRF.
The table shows that both methods perform significantly worse than the real dataset.
This proves that there is a significant gap between real blur and synthetic blur generated by the na\"{i}ve averaging approach of the previous synthetic dataset.
The table also shows that considering CRF is important for the deblurring performance of real blurred images.

\paragraph{Frame Interpolation (3, 4, 5 \& 6)}
We then study the effect of frame interpolation, which is used to fill the temporal gap between consecutive sharp frames by the REDS~\cite{Nah_NTIRE19} and DVD~\cite{Su-CVPR17} datasets.
Methods 5 and 6 in \Tbl{analysis_result} use frame interpolation.
The method 6 adds synthetic Gaussian noise to its images as described in \Sec{blur_synthesis}.
Interestingly, the table shows that the method 5 performs worse than the method 3 without frame interpolation.
This is because of the different amounts of noise in blurred images of methods 3 and 5.
As frame interpolation increases the number of frames, more frames are averaged to produce a blurred image. Thus, a resulting blurred image has much less noise.
The results of methods 6 and 4, both of which add Gaussian noise, verify this.
The results show that frame interpolation performs better than na\"{i}ve averaging when Gaussian noise is added.

\paragraph{Saturation (6, 7, 8, 9 \& 10)}
To analyze the effect of saturated pixels, we first compare the method 6, which does not include saturated pixels whose values are clipped, and the method 7, which uses oracle saturated pixels.
As shown by the results, including saturated pixels improves the deblurring quality by 0.12 dB.
Especially, the improvement is large for the test images with saturated pixels (0.30 dB).
Methods 8 and 10 use our saturation synthesis approach.
The result of the method 8 shows that, while it is worse than the method 7 (the oracle method), it still performs better the method 6, which does not perform saturation synthesis, especially for the test images with saturated pixels. 
Also, our final method (method 10) performs comparably to the oracle method (method 9).
Both methods 9 and 10 achieve 32.06 dB for all the test images.
This confirms the effectiveness of our saturation synthesis approach despite its simplicity. 

\paragraph{Noise \& ISP (5, 6, 7, 8, 9 \& 10)}
We study the effect of noise and the ISP.
To this end, we compare three different approaches:
1) ignoring noise, 2) adding Gaussian noise, and 3) adding Gaussian and Poisson noise with an ISP. The first approach corresponds to previous synthetic datasets that do not consider noise, such as GoPro~\cite{Nah_2017_CVPR}, REDS~\cite{Nah_NTIRE19} and DVD~\cite{Su-CVPR17}.
The second is the most widely used approach for generating synthetic noise in many image restoration tasks~\cite{zhang2017beyond}.
The third one reflects real noise and distortion caused by an ISP.

The table shows that, compared to the method 5 (No noise), the method 6 (Gaussian noise) performs significantly better by 1.57 dB.
Moreover, a comparison between methods 7 and 8 (Gaussian noise) and methods 9 and 10 (Gaussian+Poisson noise with an ISP) shows that adding more realistic noise and distortion further improves the deblurring performance consistently.

Finally, our final method (method 10) achieves 32.06 dB, which is more than 1 dB higher than those of the na\"{i}ve methods 2, 3, and 5. 
In terms of SSIM, our final method outperforms the na\"{i}ve methods by more than 0.05.
Compared to all the other methods, our final method achieves the smallest difference against the method 1 which uses real blurred images.
In terms of SSIM, our final method achieves 0.8332, which is only 0.0076 lower than that of the method 1.
This proves the effectiveness of our method, and the importance of realistic blur synthesis.

\paragraph{Qualitative Examples}
\Fig{qualitative_results}(b)-(e) show qualitative deblurring results produced by models trained with different methods in \Tbl{analysis_result}.
As \Fig{qualitative_results}(c) shows, a deblurring model trained with images synthesized using frame interpolation without noise synthesis fails to remove blur in the input blurred image.
Adding Gaussian noise improves the quality (\Fig{qualitative_results}(d)), but blur still remains around the lights. Meanwhile, the method trained with our full pipeline (\Fig{qualitative_results}(e)) produces a comparable result to the method trained with real blurred images.

\vspace{-5pt}
\subsection{Application to Other Datasets}
\label{sec:application_to_other_dataset}

\afterpage{
\setlength{\tabcolsep}{4pt}
\begin{table*}[t]
\centering
\vspace{-0.3cm}
\caption{ Performance comparison of different blur synthesis methods on the RealBlur\_J~\cite{jsrim-ECCV2020} and BSD\_All~\cite{Zhong_2020_ECCV,Zhong_2021_arxiv} test sets. Interp.: Frame interpolation. Sat.: Saturation synthesis. sRGB: Gamma correction of sRGB space. G: Gaussian noise. G+P: Gaussian and Poisson noise. A7R3: Using camera ISP parameters estimated from a Sony A7R3 camera, which was used for collecting the RealBlur dataset.}
\label{tbl:other_dataset_result}
\vspace{-0.2cm}
\scalebox{0.90}{
\begin{tabular}{|ccccccc|cc|}
\hline
\multicolumn{7}{|c|}{Blur Synthesis Methods}                                                         & \multicolumn{2}{c|}{PSNR / SSIM}       \\ \hline
No. & Training set & CRF    & Interp.             & Sat. & Noise         & ISP             & RealBlur\_J    & BSD\_All              \\ \hline
1 & RealBlur\_J  &        &                           &            &               &                  & 30.79 / 0.8985 & 29.67 / 0.8922        \\
2 & BSD\_All     &        &                           &            &               &                 & 28.66 / 0.8589 & 33.35 / 0.9348        \\
3 & GoPro     & Linear &                           &            &               &                  & 28.79 / 0.8741 & 29.17 / 0.8824        \\
4 & GoPro     & sRGB    &                           &            &               &                  & 28.93 / 0.8738 & 29.65 / 0.8862        \\
5 & GoPro      & sRGB    & \checkmark &            &               &                 & 28.92 / 0.8711 & 30.09 / 0.8858        \\
6 & GoPro      & sRGB    & \checkmark &            & G             &                 & 29.17 / 0.8795 & 31.19 / 0.9147        \\
7 & GoPro      & sRGB    & \checkmark & Ours    & G             &                 & 29.95 / 0.8865 & 31.41 / 0.9154        \\
8 & GoPro      & sRGB, A7R3    & \checkmark & Ours   & G+P          & A7R3 & 30.32 / 0.8899 & 30.48 / 0.9060 \\ 
\hline
9 & GoPro\_U     & Linear &                           &            &               &                  & 29.09 / 0.8810 & 29.22 / 0.8729        \\
10 & GoPro\_U     & sRGB    &                           &            &               &                  & 29.28 / 0.8766 & 29.72 / 0.8773        \\
11 & GoPro\_U     & sRGB    &                           &            & G &                  & 29.50 / 0.8865 & 30.22 / 0.8973        \\
12 & GoPro\_U     & sRGB    &                           & Ours & G  &                  & 30.40 / 0.8970 & 30.31 / 0.8995        \\
13 & GoPro\_U     & sRGB, A7R3    &                           &  Ours & G+P  & A7R3 & 30.75 / 0.9019 & 29.72 / 0.8925        \\
\hline
\end{tabular}
}
\vspace{-0.35cm}
\end{table*}

\begin{figure}[t]
\centering
\includegraphics[width=1\linewidth]{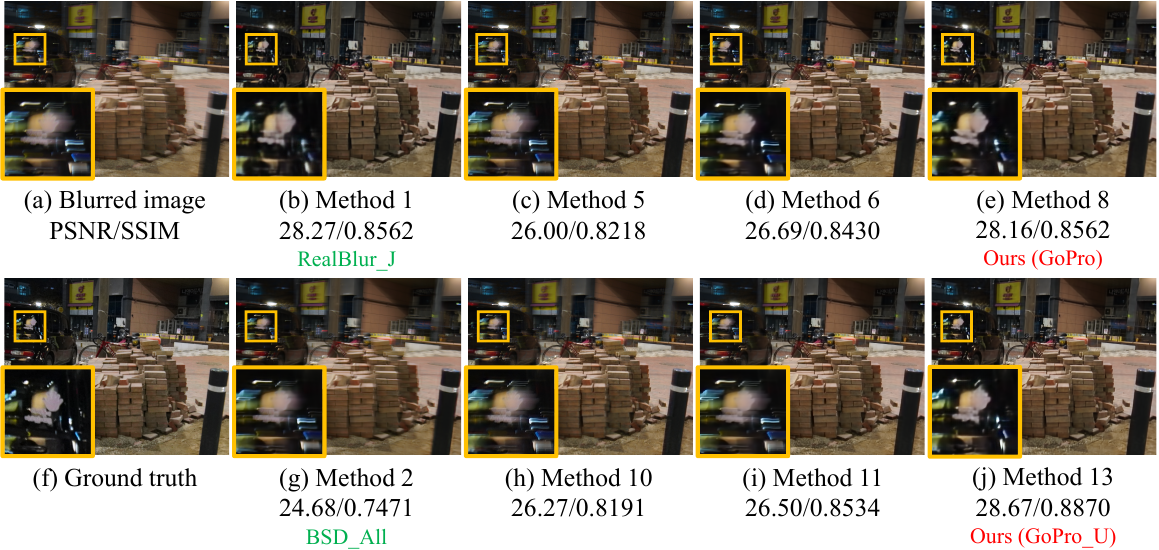}
\vspace{-0.8cm}
\caption{Qualitative comparison of deblurring results on the RealBlur\_J test set produced by models trained with different synthesis methods.
(b)-(e) Methods 1, 5, 6 and 8 in \Tbl{other_dataset_result}.
(g)-(j) Methods 2, 10, 11 and 13 in \Tbl{other_dataset_result}.
Best viewed in zoom in.
}
\vspace{-0.50cm}
\label{fig:qualitative_results_reablur}
\end{figure}
}

We evaluate the proposed pipeline on other datasets.
Specifically, we apply several variants of our pipeline to the sharp source images of the GoPro dataset~\cite{Nah_2017_CVPR} to synthesize more realistic blurred images.
Then, we train a deblurring model on synthesized images, and evaluate its performance on the RealBlur\_J~\cite{jsrim-ECCV2020} and BSD~\cite{Zhong_2020_ECCV,Zhong_2021_arxiv} datasets.
The BSD dataset consists of three subsets with different shutter speeds.
We use all of them as a single set, which we denote by BSD\_All.

\paragraph{Limited Coverage of Real Datasets}
We examine the performance of real datasets on other real datasets to study the coverage of real datasets.
To this end, we compare methods 1 and 2 in \Tbl{other_dataset_result}. The comparison shows that the performance of a deblurring model on one dataset significantly drops when trained on the other dataset.
This proves the limitation of the existing real datasets and the need for a blur synthesis approach that can generate realistic datasets for different camera settings.

\paragraph{Improving GoPro}
The method 3 in \Tbl{other_dataset_result} performs na\"{i}ve averaging to the sharp source images in the GoPro dataset~\cite{Nah_2017_CVPR} without gamma correction.
The method 4 performs gamma decoding, na\"{i}ve averaging, and then gamma encoding.
These two methods correspond to the original generation processes of GoPro. 
As the images in both RealBlur\_J~\cite{jsrim-ECCV2020} and BSD~\cite{Zhong_2020_ECCV,Zhong_2021_arxiv} datasets have blur distorted by the CRFs, the method 4 performs better. However, both of them perform much worse than the real-world blur training sets for both RealBlur\_J and BSD\_All.

The method 5 performs 0.01 dB worse than the method 4 on RealBlur\_J. 
This again shows that frame interpolation without considering noise may degrade the deblurring performance as it reduces noise as discussed in \Sec{analysis_using_RSBlur}.
Adding Gaussian noise (method 6), and saturated pixels (method 7) further improves the deblurring performance on both test sets.
For Gaussian noise, we simply add Gaussian noise with standard deviation $\sigma=0.0112$.

The method 8 uses the noise and ISP parameters estimated for the RealBlur\_J dataset~\cite{jsrim-ECCV2020}.
The RealBlur\_J dataset was captured using a Sony A7R3 camera, of which we can estimate the noise distribution, color correction matrix, and CRF.
We use the method described in \Sec{blur_synthesis} for noise estimation. For the color correction matrix and CRF estimation, we refer the readers to our supplementary material.
As the CRFs of the training set and RealBlur\_J are different, the method 8 uses different CRFs in different steps.
Specifically, it uses gamma decoding with sharp source images of the GoPro dataset into the linear space, and in the last step of our pipeline, it applies the estimated CRF of Sony A7R3.
The method 8 achieves 30.32 dB for the RealBlur\_J dataset, which is much higher than 28.93 dB of the method 4.
This proves that our blur synthesis pipeline reflecting the noise distribution and distortion caused by the ISP improves the quality of synthetic blur.
It is also worth mentioning that the method 8 performs worse than the method 7 on the BSD\_All dataset because the camera ISP of BSD\_All is different from that of RealBlur\_J.
This also shows the importance of correct camera ISP parameters including CRFs, and explains why real datasets perform poorly on other real datasets.
\Fig{qualitative_results_reablur}(c)-(e) show results of deblurring methods trained on the GoPro dataset with different methods in \Tbl{other_dataset_result}.

\paragraph{Convolution-Based Blur Synthesis}
Our pipeline also applies to convolution-based blur synthesis and improves its performance as well.
To verify this, we build a dataset with synthetic blur kernels as follows.
For each sharp image in the GoPro dataset, we randomly generate ten synthetic blur kernels following \cite{jsrim-ECCV2020} in order that we can convolve them with sharp images to synthesize blurred images instead of frame interpolation and averaging.
We also compute saturation masks $M_{sat}$ by convolving the masks of saturated pixels in each sharp image with the synthetic blur kernels.
We denote this dataset as GoPro\_U.
Methods 9 to 13 in \Tbl{other_dataset_result} show different variants of our pipeline using GoPro\_U.
For these methods, except for the frame interpolation and averaging, all the other steps in our pipeline are applied in the same manner.

In \Tbl{other_dataset_result},
methods 9 and 10 perform much worse than methods 1 and 2, which use real blurred images.
Methods 11 and 12 show that considering Gaussian noise ($\sigma=0.0112$) and saturated pixels significantly improves the performance.
Finally, the method 13 that uses our full pipeline, achieves 30.75 dB in PSNR, which is only 0.04 dB lower than that of the method 1, and achieves 0.9019 in SSIM, which is 0.003 higher than that of the method 1.
This shows that our pipeline is also effective for the convolution-based blur model.
\Fig{qualitative_results_reablur}(h)-(j) show results of deblurring methods trained on the GoPro\_U dataset with different methods in \Tbl{other_dataset_result}. 

%% file: 5.conclusion.tex
\vspace{-5pt}
\section{Conclusion}
In this paper, we presented the RSBlur dataset, which is the first dataset that provides pairs of a real-blurred image and a sequence of sharp images.
Our dataset enables accurate analysis of the difference between real and synthetic blur.
We analyzed several factors that introduce the difference 
between them with the dataset and presented a novel blur synthesis pipeline, which is simple but effective.
Using our pipeline, we quantitatively and qualitatively analyzed the effect of each factor that degrades the deblurring performance on real-world blurred images.
We also showed that our blur synthesis pipeline could greatly improve the deblurring performance on real-world blurred images.

\paragraph{Limitations and Future Work}
Our method consists of simple and heuristic steps including a simple ISP and a mask-based saturation synthesis. While they improve the deblurring performance, further gains could be obtained by adopting sophisticated methods for each step.
Also, there is still the performance gap between real blur datasets and our synthesized datasets, and some other factors may exist that cause the gap.
Investigating that is an interesting future direction.

%% file: eccv2022_supplementary.tex
\title{Realistic Blur Synthesis for Learning Image Deblurring \\
—– Supplementary Material —–} 

\titlerunning{Realistic Blur Synthesis for Learning Image Deblurring}
%
\author{Jaesung Rim \and
Geonung Kim \and
Jungeon Kim \and
Junyong Lee \and \\
Seungyong Lee \and
Sunghyun Cho}
\authorrunning{J. Rim et al.}
%
\institute{
POSTECH, Pohang, Korea \\
\email{\{jsrim123,k2woong92,jungeonkim,junyonglee,leesy,s.cho\}@postech.ac.kr}}
\maketitle

\section{Statistics of the RSBlur}

\Tbl{statistical_reports} shows a statistical comparison with the RSBlur and other real-world blur datasets.
The proposed RSBlur dataset consists of 13,358 real blurred images, which make the dataset the second largest real-world dataset.
While the RealBlur~\cite{jsrim-ECCV2020} and BSD~\cite{Zhong_2020_ECCV,Zhong_2021_arxiv} datasets consist of real blurred images and ground-truth sharp images, we provide real blurred images and sequences of nine sharp images to enable analysis on the blur generation process between real blurred and synthetic blurred images. In terms of image resolution, the RSBlur dataset provides the largest resolution.
We also report the estimated noise levels using a single image noise estimation method~\cite{Chen-ICCV15} for comparing the amounts of noise in the real-world blurred datasets.

\begin{table}[h]
\centering
\setlength{\tabcolsep}{3.0pt}
\caption{Statistical comparison of real-world blur datasets. The average noise levels are estimated using a single image noise estimation method~\cite{Chen-ICCV15}.}
\label{tbl:statistical_reports}
\scalebox{1.0}{
\begin{tabular}{|cccccc|}
\hline
         & Frames & Real/Synth.         & Resolution       & Shutter (ms) & Noise \\ \hline
RealBlur~\cite{jsrim-ECCV2020} & 4,738    & Real & $680\times773$   & 500          & 0.4378 \\
BSD~\cite{Zhong_2020_ECCV,Zhong_2021_arxiv}      & 33,000   & Real       & $640\times480$   & 8, 16, 24    & 0.3404 \\ 
RSBlur   & 13,358   & Real \& Synth       & $1920\times1200$ & 100          & 0.7736 \\ \hline
\end{tabular}
}
\end{table}

\section{Real-world Deblurring Benchmark on the RSBlur}
While there exist a couple of real-world blur datasets such as RealBlur~\cite{jsrim-ECCV2020} and BSD~\cite{Zhong_2020_ECCV,Zhong_2021_arxiv},
their coverage is limited.
The real-world blurred images in the RSBlur dataset can also serve as an additional benchmark dataset that complements the existing benchmark datasets in terms of coverage.

In this section, we provide a benchmark on recent state-of-the-art deblurring methods using the RSBlur dataset to provide a basis for future deblurring research.
Using real-world blurred images of the RSBlur dataset, we train state-of-the-art deblurring methods~\cite{Tao-CVPR18,Cho_2021_ICCV,Zamir_2021_CVPR,Zamir_2022_CVPR,Wang_2022_CVPR}.
We use the source codes provided by the authors for training, and evaluate their performance using the real-blur test set of the RSBlur dataset.
Here, we briefly report the qualitative and quantitative results of the state-of-the-art methods in \Tbl{deblurring_benchmark} and in \Fig{our_system_diagram}, respectively.

\setlength{\tabcolsep}{4pt}
\begin{table}[t]
\caption{Benchmark of state-of-the-art deblurring methods on the real blurred test set of the RSBlur dataset. We trained all methods using real blurred training set of the RSBlur dataset.}
\label{tbl:deblurring_benchmark}
\centering
\scalebox{1.0}{
\begin{tabular}{|cc|}
\hline
\multicolumn{1}{|c|}{Methods}    & PSNR / SSIM      \\ \hline
\multicolumn{1}{|c|}{SRN-Deblur~\cite{Tao-CVPR18}} & 32.53 / 0.8398 \\
\multicolumn{1}{|c|}{MiMO-UNet~\cite{Cho_2021_ICCV}}  & 32.73 / 0.8457 \\
\multicolumn{1}{|c|}{MiMO-UNet+~\cite{Cho_2021_ICCV}} & 33.37 / 0.8560 \\
\multicolumn{1}{|c|}{MPRNet~\cite{Zamir_2021_CVPR}}     & 33.61 / 0.8614 \\
\multicolumn{1}{|c|}{Restormer~\cite{Zamir_2022_CVPR}}  & 33.69 / 0.8628 \\ 
\multicolumn{1}{|c|}{Uformer-B~\cite{Wang_2022_CVPR}}    & 33.98 / 0.8660 \\ \hline
\end{tabular}
}
\end{table}

\begin{figure}[t]
\centering
\includegraphics[width=0.95\linewidth]{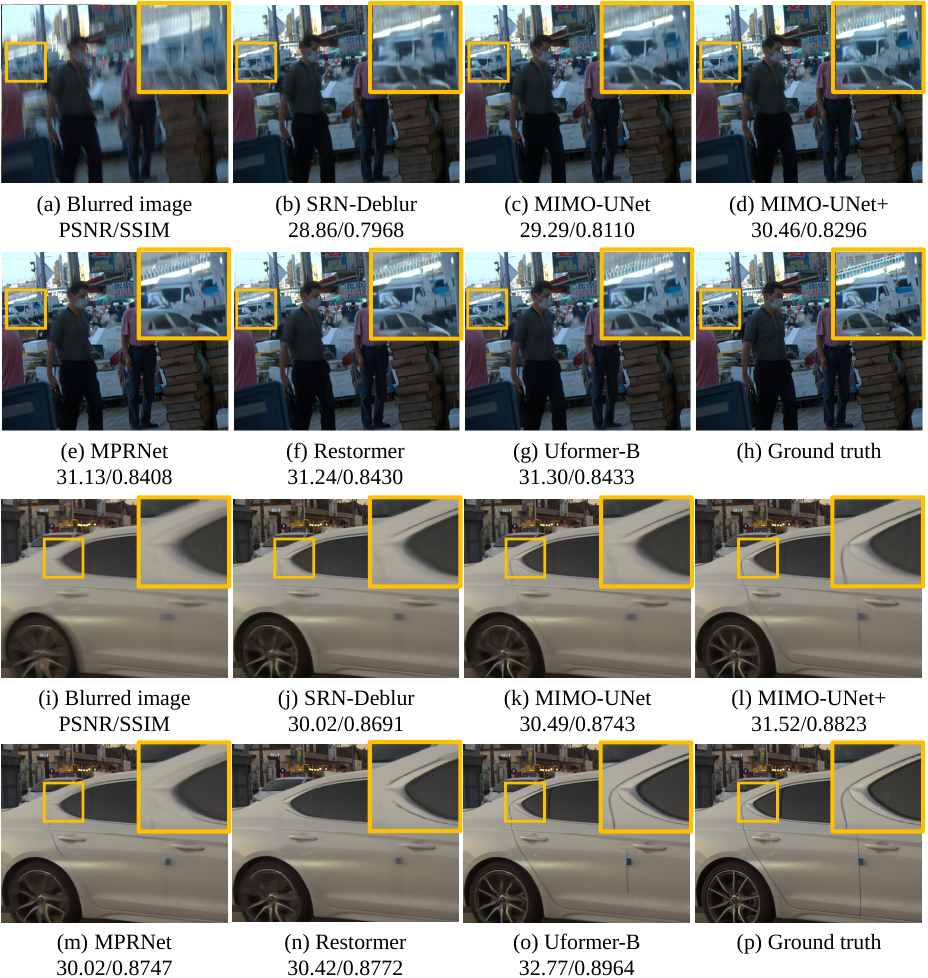}
\caption{Qualitative comparison of state-of-the-art deblurring methods on real-world blurred images of the RSBlur test set.}
\label{fig:RSBlur_benchmark_results}
\end{figure}

\section{Details of Dual-Camera System}

\begin{figure}[h]
\begin{center}
\includegraphics [width=0.95\linewidth] {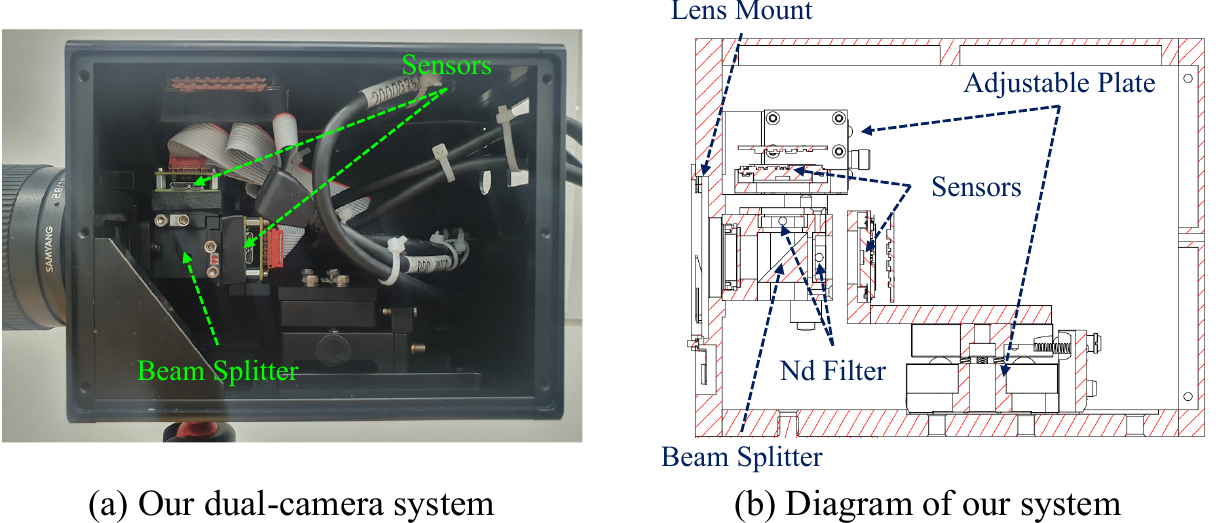}
\end{center}
\vspace{-0.5cm}
\caption{The dual-camera system and detailed diagram of the system.}
\label{fig:our_system_diagram}
\vspace{-0.6cm}
\end{figure}

In this section, we describe the details of our dual-camera system. \Fig{our_system_diagram} shows our dual-camera system and a diagram of the system.  The system consists of a mount for the lens, one beam splitter, and two camera modules with imaging sensors (Basler daA1920-160uc) so that the camera modules can capture the same scene while sharing one lens.
For the lens, we used a Samyang 10mm F2.8 ED AS NCS CS.
We installed a 5\% neutral density filter (OD 1.3 VIS, 12.5mm Dia. Non-Reflective ND Filter) in front of a camera module. For compensation of beam-splitter tolerance, a 63\% neutral density filter (0.2 OD, 25mm Dia., Precision Absorptive ND Filter) is installed in front of the other camera module. Two camera modules are installed on adjustable plates, so we can physically align the modules by adjusting the plates.

One camera module with a 5\% neutral density filter captures a blurred image with a long exposure time (0.1 seconds). The other module captures nine sharp images with a short exposure time (0.005 seconds) during the exposure time of a blurred image. The gains of the two modules are set to 0 for both. To increase the number of images and diversity of blur, we capture 20 pairs of a blurred image and a sequence of nine sharp images of the same scene.

\section{Camera ISP}
\label{sec:details_ISP}

To collect our dataset, we captured all images in the camera RAW format, and converted them into the nonlinear sRGB space using a simple image signal processing (ISP) pipeline.
Our ISP consists of four steps: 1) white balance, 2) demosaicing, 3) color correction, and 4) conversion to the sRGB space using a camera response function.
We use the demosaicing method of Malvar \etal~\cite{Malvar_2004_ICASSP} for the second step and a gamma correction of standard RGB space for the fourth step.

For the white balance and color correction steps in our ISP, we utilize a color chart.
Specifically, when collecting our dataset, we captured reference images of a color chart for different scenes.
Using the reference images, we estimate the gain $g_c$ for the color channel $c \in \{R,G,B\}$ for the white balance as:
\begin{equation}
    g_c = \frac{P_{n}(G)}{P_{n}(c)} \label{eq:gain}
\end{equation}
where $P_{n}(c)$ is the mean intensity of the color channel $c$ of neutral patches in the color chart.
Then, in the first step of our ISP, each color channel of a RAW image is multiplied by the corresponding gain.

For the color correction in the third step of our ISP,
we estimate a color correction matrix of the XYZ color space as:
\begin{eqnarray}
\begin{bmatrix}
X_1^{ref} & \cdots  & X_{24}^{ref} \\ 
Y_1^{ref} & \cdots  & Y_{24}^{ref} \\ 
Z_1^{ref} & \cdots  & Z_{24}^{ref}
\end{bmatrix} = \alpha \cdot T \begin{bmatrix}
X_1 & \cdots  & X_{24} \\ 
Y_1 & \cdots  & Y_{24} \\ 
Z_1 & \cdots  & Z_{24}
\end{bmatrix}
\label{eq:color_correction_matrix}
\end{eqnarray}
where $(X_i^{ref}, Y_i^{ref}, Z_i^{ref})$ is the reference XYZ color of the $i$-th color chart patch,
and $(X_i, Y_i, Z_i)$ is the measured XYZ color after white balancing and demosaicing.
$\alpha$ is a single scalar value for matching the brightness levels of the color chart patches and the captured patches.
$T$ is a $3\times3$ color correction matrix.
We first estimate $\alpha$ by minimizing the mean-squared error between the color chart patches and captured patches.
Then, we estimate $T$ by finding the least-squares solution of \Eq{color_correction_matrix} with fixed $\alpha$.

\begin{figure}[t]
\begin{center}
\includegraphics [width=1\linewidth] {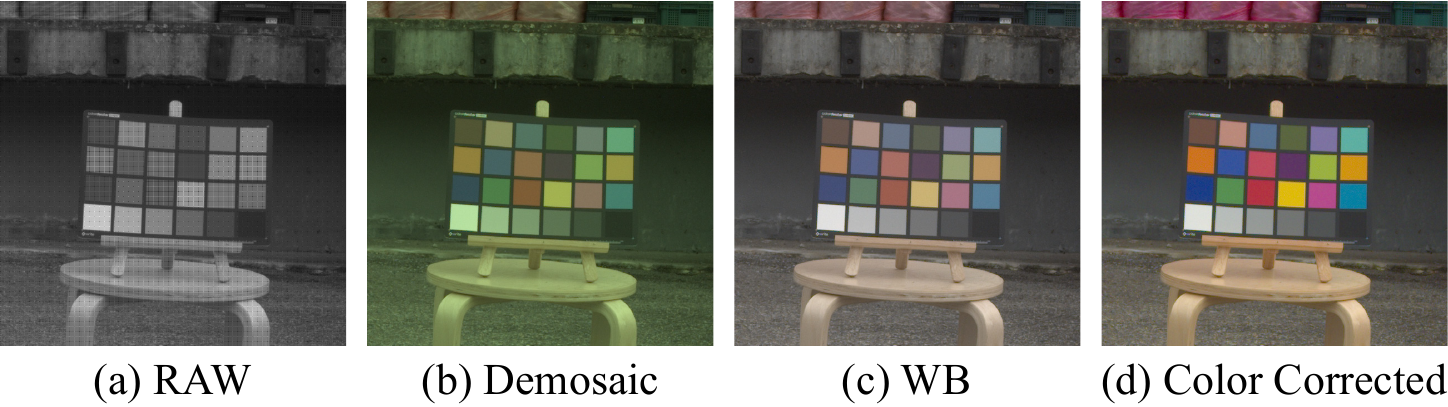}
\end{center}
\vspace{-0.4cm}
\caption{Intermediate images of our camera ISP. All images are gamma corrected for visualization. }
\label{fig:example_isp}
\end{figure}

Once a color correction matrix $T$ is obtained, we apply $T$ in the third step of our ISP as follows:
\begin{equation}
    \begin{bmatrix} R_{Lin} \\ G_{Lin} \\ B_{Lin} \end{bmatrix} = M_{XYZ2Lin} \cdot T \cdot M_{Lin2XYZ} \cdot \begin{bmatrix} R_{Dem} \\ G_{Dem} \\ B_{Dem} \end{bmatrix}  \label{eq:cam2lin}
\end{equation}
where $(R_{Dem}, G_{Dem}, B_{Dem})$ is an RGB color of an image after demosaicing, and $(R_{Lin}, G_{Lin}, B_{Lin})$ is a resulting RGB color in the linear sRGB space.
$M_{Lin2XYZ}$ and $M_{XYZ2Lin}$ are matrices for color conversion between the linear sRGB and XYZ color spaces.
\Fig{example_isp} shows intermediate results of our camera ISP.

\section{Photometric Alignment between Camera Modules}

\begin{figure}[t]
\begin{center}
\includegraphics [width=1\linewidth] {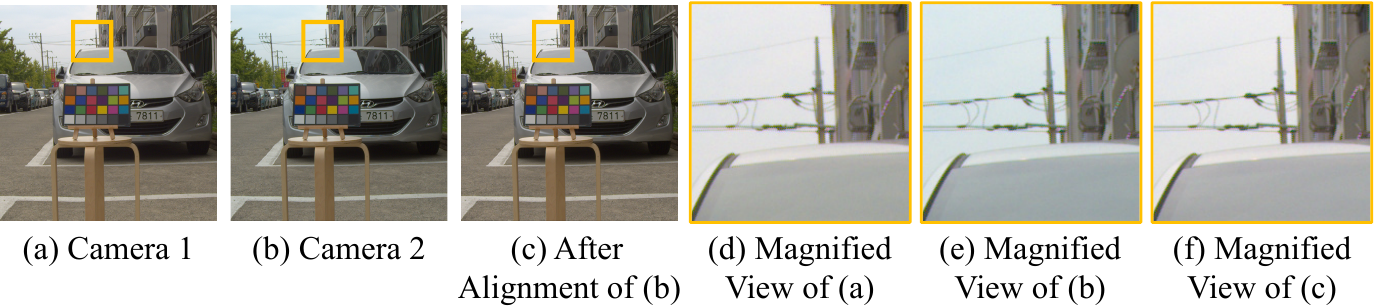}
\end{center}
\vspace{-0.4cm}
\caption{Result of photometric alignment. (a) \& (b) show captured images from a one camera module and the other camera module, respectively. (c) shows the photometric alignment result of (b). (d)-(f) show magnified views of (a)-(c).}
\label{fig:photometric_alignment}
\end{figure}

Due to the optical spectrum difference caused by the beam splitter and ND filters, captured images may have slight photometric misalignments. To mitigate this, we conduct photometric alignment using a color chart image after the color correction step of our camera ISP.
Specifically, we formulate the relationship between the colors of images from the two camera modules as:

\begin{eqnarray}
\begin{bmatrix}
X_1^{C1} & \cdots  & X_{24}^{C1} \\ 
Y_1^{C1} & \cdots  & Y_{24}^{C1} \\ 
Z_1^{C1} & \cdots  & Z_{24}^{C1}
\end{bmatrix} = T_{p} \begin{bmatrix}
X_1^{C2} & \cdots  & X_{24}^{C2} \\ 
Y_1^{C2} & \cdots  & Y_{24}^{C2} \\ 
Z_1^{C2} & \cdots  & Z_{24}^{C2}
\end{bmatrix}
\label{eq:photometric_alignment}
\end{eqnarray}
where $(X_i^{C1}, Y_i^{C1}, Z_i^{C1})$ and $(X_i^{C2}, Y_i^{C2}, Z_i^{C2})$ are the XYZ color values of the $i$-th color chart patch after the color correction of one camera module (C1) and the other camera module (C2), respectively. $T_{p}$ is a $3\times3$ matrix for the photometric alignment. We estimate $T_{p}$ by finding the least-squares solution of \Eq{photometric_alignment}. 
As shown in \Fig{photometric_alignment}(a)-(b), images captured by the two camera modules have color differences. After photometric alignment, the color difference is significantly reduced, as shown in \Fig{photometric_alignment}(c).

\section{Geometric Alignment between Camera Modules}

\begin{figure}[t]
\begin{center}
\includegraphics [width=1\linewidth] {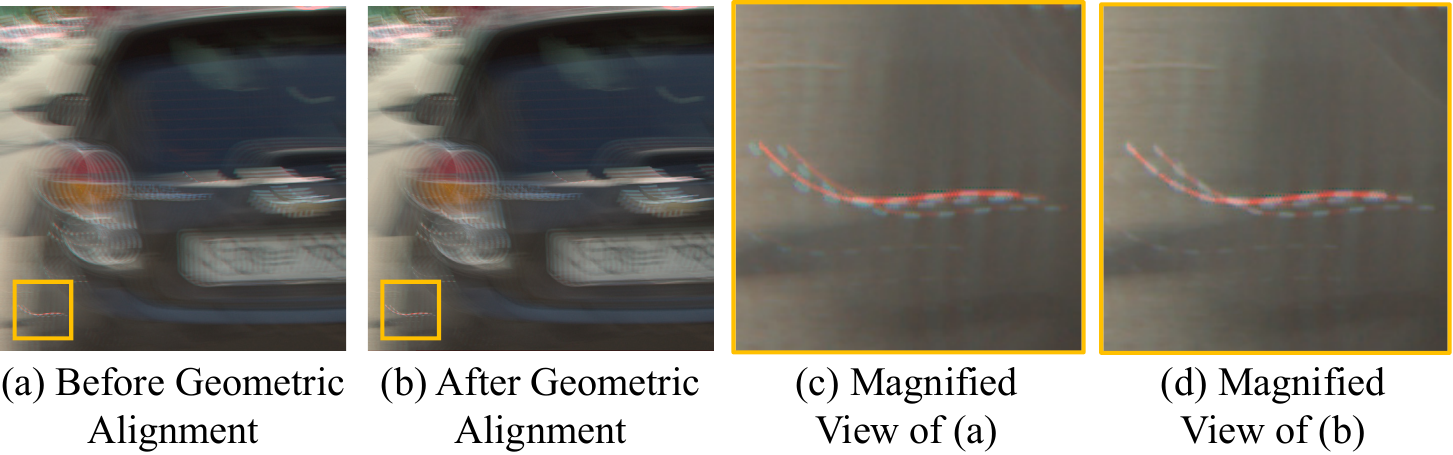}
\end{center}
\vspace{-0.4cm}
\caption{Result of geometric alignment. (a) \& (b) show stereo-anaglyph images, where a real blurred and the averaging of nine sharp images are visualized in red and cyan, respectively. (c) \& (d) show magnified views of (a) and (b).}
\label{fig:geometric_alignment}
\end{figure}

Although the two camera modules are physically aligned as much as possible, there may exist a small amount of geometric misalignment between images from them (\Fig{geometric_alignment}(c)).
Thus, after capturing images, we conduct geometric alignment to compensate for this.
The geometric alignment is performed for each pair of a blurred image and its corresponding sharp image sequence as the degree of misalignment can vary with respect to the distance between the camera system and scene.

Specifically, for a given sharp image sequence, we first increase the frame rate 8$\times$ using a frame interpolation method~\cite{Park_2021_ICCV}, and synthesize a blurred image by averaging them.
Then, we estimate a homography between a real blurred image and the synthesized one using the enhanced correlation coefficient method~\cite{Evangelidis-TPAMI08}.
Finally, we warp the sharp images according to the estimated homography.
We perform geometric alignment to the images processed by the ISP.
\Fig{geometric_alignment}(d) shows a result of our geometric alignment, where the red and cyan lights are better aligned after geometric alignment. Also, it shows that the real blurred image and nine sharp images are well synchronized.

\section{Camera ISP for the RealBlur Dataset} 
\label{sec:details_ISP_A7R3}

In the main paper, we improve the performance of SRN-DeblurNet~\cite{Tao-CVPR18} by mimicking the ISP of the Sony A7R3 camera, which is used for the RealBlur dataset~\cite{jsrim-ECCV2020}. 
Our camera ISP for the Sony A7R3 also consists of the white balance, demosaicing, color correction, and camera response function (CRF) steps. 

As the ISP affects the noise distribution and non-linearity of the blur, we match the white balance gains, color correction matrix, and CRF as much as possible to those of the RealBlur dataset. For white balance, we extract the white balance gains from the RAW images of the RealBlur training set.
Similar to~\cite{Brooks_2019_CVPR_denoising}, we randomly sample the gains from the RealBlur training set and multiply a RAW image by the gains. After white balancing, we apply the demosaicing method of Malvar \etal~\cite{Malvar_2004_ICASSP}. 

For the color correction, we extract a characterization matrix of A7R3 from the Libraw library and convert it into a color correction matrix following~\cite{Andrew_2020_Optical_Engineering}. 
The extracted color correction matrix directly maps from the RAW space to the XYZ color space.
We convert a demosaiced image into the linear sRGB space as:
\begin{equation}
    \begin{bmatrix} R_{Lin} \\ G_{Lin} \\ B_{Lin} \end{bmatrix} = M_{XYZ2Lin} \cdot T_{A7R3} \cdot \begin{bmatrix} R_{Dem} \\ G_{Dem} \\ B_{Dem} \end{bmatrix}  \label{eq:cam2lin_a7r3}
\end{equation}
where $(R_{Dem}, G_{Dem}, B_{Dem})$ is an RGB color of an image after demosaicing, and $(R_{Lin}, G_{Lin}, B_{Lin})$ is a resulting RGB color in the linear sRGB space.

\begin{figure}[t]
\begin{center}
\includegraphics [width=1.0\linewidth] {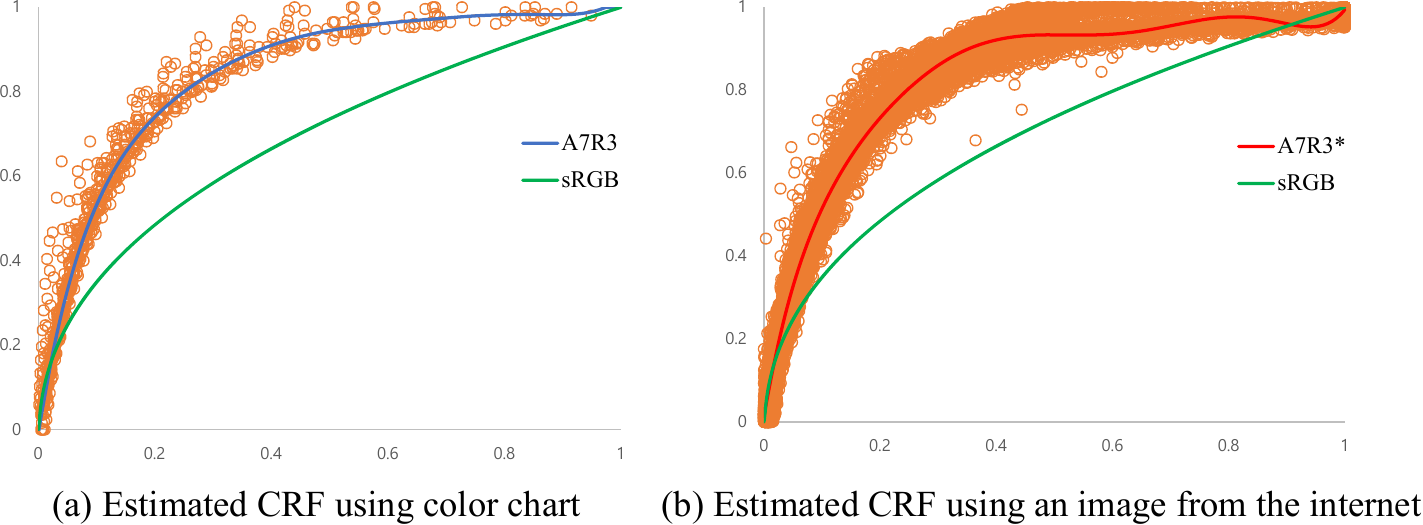}
\end{center}
\vspace{-0.4cm}
\caption{The blue and red lines show the estimated CRFs of the Sony A7R3 using color chart images and a raw-RGB image from the internet, respectively. The orange dots show measured values. The green lines show gamma correction of sRGB space.}
\label{fig:A7R3_CRF}
\end{figure}

The final step applies a CRF. Motivated by~\cite{Mitsunaga_1999_CVPR}, we model the CRF as a high-order polynomial as follows:
\begin{equation}
    I_{srgb} = \sum_{k=0}^{K} c_k I_{Lin}^k \label{eq:crf}
\end{equation}
where $I_{srgb}$, $I_{Lin}$, and $c_k$ are a non-linear and a linear sRGB image, and polynomial coefficients, respectively. $K$ is the polynomial order, which is set to 7. We capture 11 color chart images with different shutter speeds using a Sony A7R3 camera.
Then, we measure the RGB values of the color patches in JPEG images from the camera and in linear sRGB images converted from RAW images.
Using the pairs of the measured RGB values, we estimate the coefficients of the CRF $c_k$ by solving a least-squares problem.
\Fig{A7R3_CRF}(a) shows the estimated CRF using color chart images.

\begin{figure}[!t]
\centering
\includegraphics[width=0.95\linewidth]{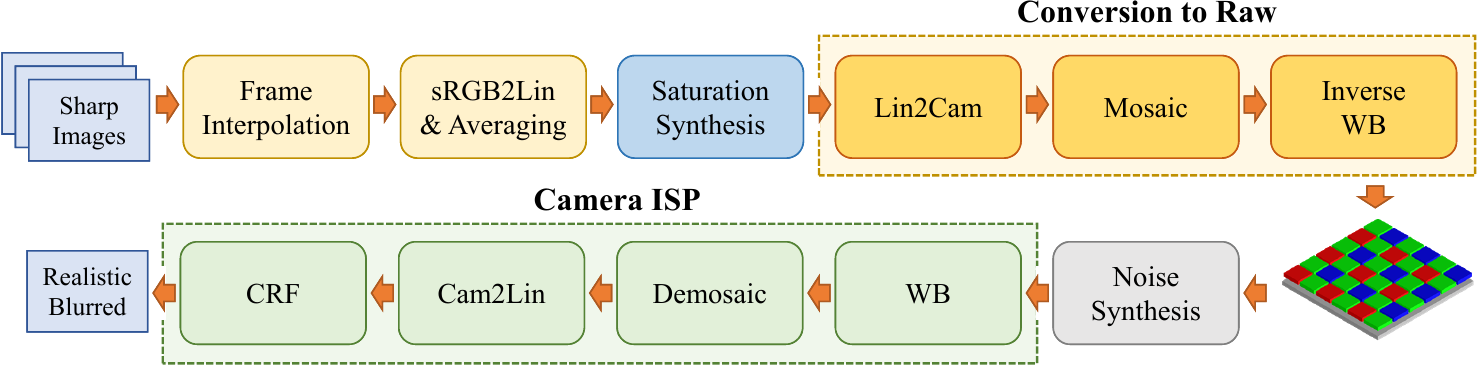}
\vspace{-0.35cm}
\caption{Overview of our realistic blur synthesis pipeline. Lin2Cam: Inverse color correction, i.e., color space conversion from the linear sRGB space to the camera RAW space. WB: White balance. Cam2Lin: Color correction.}
\vspace{-0.25cm}
\label{fig:generation_process_supple}
\end{figure}

\section{Conversion from sRGB to RAW}

\Fig{generation_process_supple} shows an overview of our realistic blur synthesis pipeline.
As described in the main paper, we convert the image from the saturation synthesis step into the mosaiced camera RAW space.
To this end, we apply inverse color correction, mosaicing, and inverse white balance sequentially.
Then, we apply the camera ISP to reflect distortions introduced by the camera ISP.
In the case of Sony A7R3, we apply the ISP described in \Sec{details_ISP_A7R3}.
In the following, we describe each step of the conversion to RAW in more detail.

\paragraph{Lin2Cam}
This step performs inverse color correction.
As color correction is a simple linear operation, we can apply inverse color correction as follows:
\begin{equation}
    \begin{bmatrix} R_{Cam} \\ G_{Cam} \\ B_{Cam} \end{bmatrix} = M_{XYZ2Lin} \cdot T^{-1} \cdot M_{Lin2XYZ} \cdot \begin{bmatrix} R_{Lin} \\ G_{Lin} \\ B_{Lin} \end{bmatrix}  \label{eq:lin2cam}
\end{equation}
where $(R_{Lin}, G_{Lin}, B_{Lin})$ and $(R_{Cam}, G_{Cam}, B_{Cam})$ are RGB colors in the linear sRGB and RAW RGB spaces, respectively.
$M_{Lin2XYZ}$ and $M_{XYZ2Lin}$ are matrices for color conversion between the linear sRGB and XYZ color spaces. $T^{-1}$ is the inverse of a color correction matrix $T$.

In the case of the Sony A7R3, the color correction matrix $T_{A7R3}$ directly maps colors in the RAW color space into the XYZ color space as mentioned in \Sec{details_ISP_A7R3}.
Thus, we perform inverse color correction as: 
\begin{equation}
    \begin{bmatrix} R_{Cam} \\ G_{Cam} \\ B_{Cam} \end{bmatrix} = T^{-1}_{A7R3} \cdot M_{Lin2XYZ} \cdot \begin{bmatrix} R_{Lin} \\ G_{Lin} \\ B_{Lin} \end{bmatrix}  \label{eq:lin2cam_a7r3}
\end{equation}
where $(R_{Lin}, G_{Lin}, B_{Lin})$ and $(R_{Cam}, G_{Cam}, B_{Cam})$ are RGB colors in the linear sRGB and RAW RGB spaces, respectively.

\paragraph{Mosaic} Following~\cite{Guo_2019_CVPR}, we randomly sample a Bayer pattern from RGGB, BGGR, GRBG, and GBRG to reflect distortions caused by various Bayer patterns.
Then, we perform mosaicing using the sampled pattern.

\paragraph{Inverse WB}
As we already know white balance gains for each image, in this step, we simply apply their inverse $g_c^{-1}$ to each color channel of a mosaiced image.

\section{Additional Analysis Results}

\afterpage{
\setlength{\tabcolsep}{3pt}
\begin{table}[t]
\centering
\caption{Additional analysis using MIMO-Unet~\cite{Cho_2021_ICCV} on the RSBlur dataset. Interp.: Frame interpolation. Sat.: Saturation synthesis. sRGB: Gamma correction of sRGB space. G: Gaussian noise. G+P: Gaussian and Poisson noise.}
\label{tbl:analysis_result_mimounet}
\vspace{-0.25cm}
\scalebox{0.9}{
\begin{tabular}{|ccccccc|c|c|c|}
\hline
\multicolumn{7}{|c|}{Blur Synthesis Methods}                                                      & \multicolumn{3}{c|}{PSNR / SSIM}                 \\ \hline
No. & Real                      & CRF    & Interp.             & Sat. & Noise & ISP     & All & Saturated      & No Saturated            \\ \hline
1 & \checkmark &        &                           &            &       &        & 32.73 / 0.8457 & 31.44 / 0.8385 & 33.93 / 0.8524  \\
2 &                           & sRGB    & \checkmark &            &       &         & 28.83 / 0.7164 & 27.42 / 0.7052 & 30.16 / 0.7270 \\
3 &                           & sRGB    & \checkmark &            & G     &         & 29.63 / 0.7552 & 28.28 / 0.7486 & 30.90 / 0.7614 \\
4 &                          & sRGB    & \checkmark & Ours    & G     &         & 29.84 / 0.7658 & 28.49 / 0.7590 & 31.12 / 0.7723 \\
5 &                          & sRGB    & \checkmark & Ours   & G+P   & \checkmark   & 32.08 / 0.8362 & 30.68 / 0.8290 & 33.39 / 0.8429 \\

\hline
\end{tabular}
}
\vspace{-0.20cm}
\end{table}

\begin{figure}[t]
\centering
\includegraphics[width=1\linewidth]{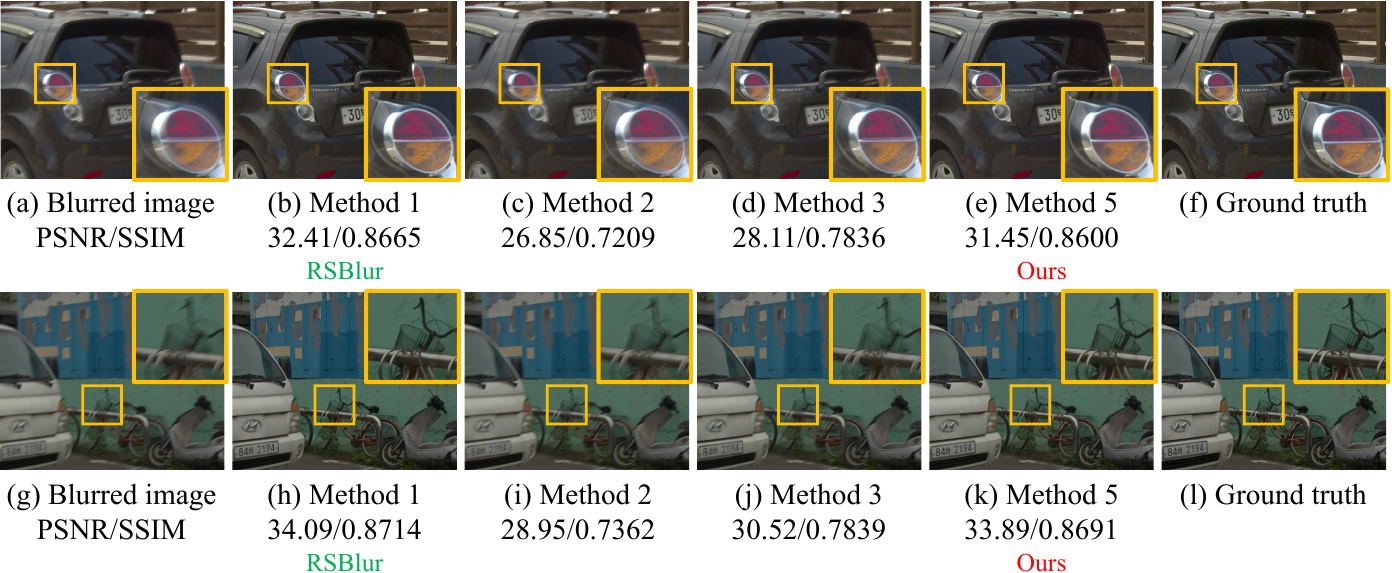}
\vspace{-0.75cm}
\caption{Qualitative comparison of deblurring results on the RSBlur test set produced by  MIMO-Unet~\cite{Cho_2021_ICCV} trained with different synthesis methods.
(b)-(e) \& (h)-(k) Methods 1, 2, 3 and 5 in \Tbl{analysis_result_mimounet}.
Best viewed in zoom in.}
\vspace{-0.3cm}
\label{fig:qualitative_results_supple}
\end{figure}
}

In this section, we also provide additional analysis results using MIMO-Unet~\cite{Cho_2021_ICCV}. For the experiments, we train MIMO-Unet model for 790K iterations, which is half the number of iterations suggested in \cite{Cho_2021_ICCV}, with variants of our pipeline.
\Tbl{analysis_result_mimounet} shows that the method 2 performs worse than method 1, which uses real blurred images for training.
Methods 3 and 4 show that adding Gaussian noise ($\sigma=0.0112$) and our saturation synthesis significantly improves the deblurring performance.
The method 5, which corresponds to our full pipeline, achieves 32.08 dB.
The analysis shows that MIMO-Unet~\cite{Cho_2021_ICCV} also has significant performance improvement with the proposed synthesis pipeline.
\Fig{qualitative_results_supple} shows results of MIMO-Unet~\cite{Cho_2021_ICCV} trained on the RSBlur dataset with different methods in \Tbl{analysis_result_mimounet}.

\section{Experiments with Rough Camera Parameters}

Estimating accurate camera-specific parameters can be easily done by taking a few shots of images.
Even if the camera is not available, rough parameters can be easily obtained using images available on the internet in the case of most consumer cameras.
To show this, assuming that an A7R3 camera is not available, we conducted additional experiments on the RealBlur\_J~\cite{jsrim-ECCV2020} dataset where we estimated camera-specific parameters from images from the internet.

In the case of most consumer cameras, the characterization matrix is easily obtained from the Libraw library\footnote{\href{https://github.com/LibRaw/LibRaw/blob/2a9a4de21ea7f5d15314da8ee5f27feebf239655/src/tables/colordata.cpp}{https://github.com/LibRaw/src/tables/colordata.cpp}}. As described in \Sec{details_ISP_A7R3}, we extract the characterization matrix of the A7R3 camera and convert it into a color correction matrix.
The SIDD dataset~\cite{Abdelhamed_2018_CVPR} provides the noise parameters of four cameras on different ISO settings. 
As the RealBlur\_J dataset is mostly captured with ISO 100, we sample the noise parameters of Google Pixel on the ISO 100 setting. 
Following~\cite{Brooks_2019_CVPR_denoising}, we randomly sample gains for the red and blue channels from $\mathcal{U}(1.9, 2.4)$ and $\mathcal{U}(1.5, 1.9)$ for the white balance.

\afterpage{
\setlength{\tabcolsep}{4pt}
\begin{table*}[t]
\centering
\caption{ Performance comparison of different blur synthesis methods on the RealBlur\_J~\cite{jsrim-ECCV2020} test sets. Interp.: Frame interpolation. Sat.: Saturation synthesis. sRGB: Gamma correction of sRGB space. G: Gaussian noise. G+P: Gaussian and Poisson noise. A7R3: Using camera ISP with accurate parameters estimated from a Sony A7R3 camera. A7R3*: Using camera ISP with rough  parameters.}
\label{tbl:rough_camera_params}
\vspace{-0.2cm}
\scalebox{0.90}{
\begin{tabular}{|ccccccc|c|}
\hline
\multicolumn{7}{|c|}{Blur Synthesis Methods}                                                         & \multicolumn{1}{c|}{PSNR / SSIM}       \\ \hline
No. & Training set & CRF    & Interp.             & Sat. & Noise         & ISP             & RealBlur\_J           \\ \hline
1 & RealBlur\_J  &        &                           &            &               &                  & 30.79 / 0.8985      \\
2 & BSD\_All     &        &                           &            &               &                 & 28.66 / 0.8589        \\
3 & GoPro      & sRGB    & \checkmark &            &               &                 & 28.92 / 0.8711        \\
4 & GoPro      & sRGB, A7R3    & \checkmark & Ours   & G+P          & A7R3 & 30.32 / 0.8899  \\ 
5 & GoPro      & sRGB, A7R3*    & \checkmark & Ours   & G+P          & A7R3* & 30.23 / 0.8864  \\ 
\hline
6 & GoPro\_U     & sRGB    &                           &            &               &                  & 29.28 / 0.8766       \\
7 & GoPro\_U     & sRGB, A7R3    &                           &  Ours & G+P  & A7R3 & 30.75 / 0.9019         \\
8 & GoPro\_U     & sRGB, A7R3*    &                           &  Ours & G+P  & A7R3* & 30.55 / 0.8956        \\
\hline
\end{tabular}
}
\vspace{-0.30cm}
\end{table*}

\begin{figure}[t]
\centering
\includegraphics[width=1\linewidth]{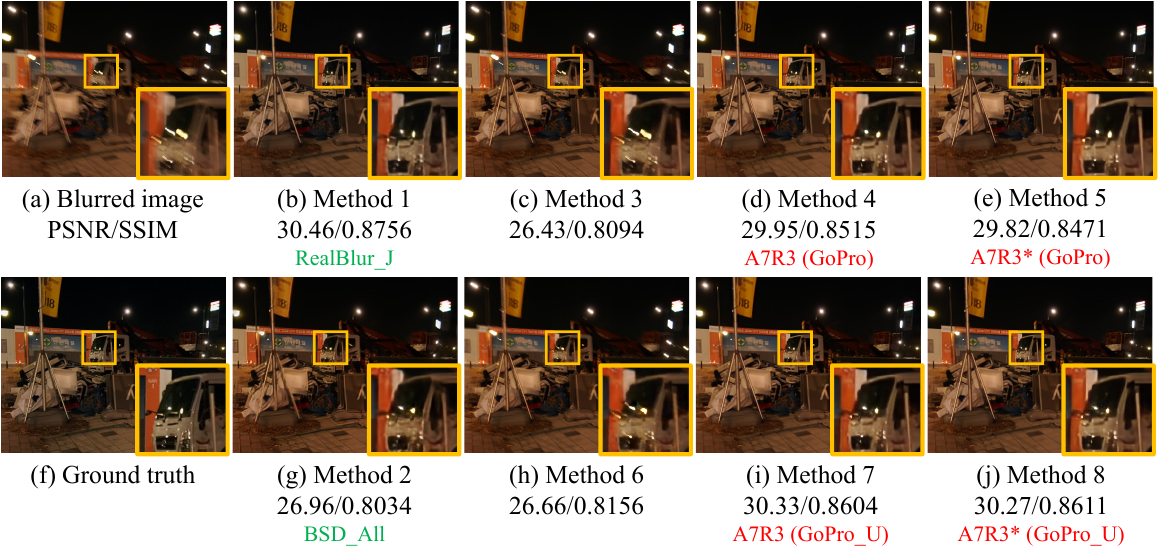}
\vspace{-0.8cm}
\caption{Qualitative comparison of deblurring results on the RealBlur\_J test set produced by models trained with different synthesis methods.
(b)-(e) Methods 1, 3, 4 and 5 in \Tbl{rough_camera_params}.
(g)-(j) Methods 2, 6, 7 and 8 in \Tbl{rough_camera_params}.
Best viewed in zoom in.
}
\label{fig:qualitative_rough_params}
\end{figure}
}

In the case of the CRF, there is no dataset including the CRFs of the latest consumer cameras, and it is difficult to estimate CRFs without cameras. Instead, we utilize the camera profile of Sony A7R3 available on Adobe Lightroom. Specifically, we first download a RAW image captured by an A7R3 from the internet. Then, we convert the RAW image into an sRGB image using the camera profile of Adobe Lightroom. We also convert the RAW RGB image into a linear sRGB image using the color correction matrix from the Libraw library and white balance gains of the RAW RGB image.
Then, using the pixel values of the converted linear sRGB image and sRGB image, we estimate the coefficients of \Eq{crf} by solving a least-squares problem.
\Fig{A7R3_CRF}(b) shows the estimated CRF using the RAW image resembles the estimated CRF using a color chart very closely.

To verify the effectiveness of the rough parameters estimated as described above, we train SRN-DeblurNet~\cite{Tao-CVPR18} using the proposed synthesis pipeline with the estimated parameters, and evaluate its performance.
In \Tbl{rough_camera_params}, methods 5 and 8 that use the rough parameters perform worse than methods 4 and 7 that use accurate camera parameters.
Neverthelss, compared to the methods 3 and 6, the methods 5 and 8 still perform significantly better, validating the effectiveness of the rough parameters.
\Fig{qualitative_rough_params} shows qualitative examples of using the proposed pipeline with rough and accurate camera parameters and other na\"{i}ve methods.

\setlength{\tabcolsep}{4pt}
\begin{table*}[t]
\centering
\caption{ Performance comparison of SRN-DeblurNet~\cite{Tao-CVPR18} trained on the RealBlur\_J~\cite{jsrim-ECCV2020}, BSD\_All~\cite{Zhong_2020_ECCV,Zhong_2021_arxiv}, and RSBlur datasets. }
\label{tbl:Limited_coverage_realdataset}
\begin{tabular}{|c|ccc|}
\hline
\multirow{2}{*}{\backslashbox{Train}{Test}} & \multicolumn{3}{c|}{PSNR / SSIM}  \\ \cline{2-4} 
                                      & \multicolumn{1}{c|}{RealBlur\_J}    & \multicolumn{1}{c|}{BSD\_All}       & RSBlur         \\ \hline
RealBlur\_J                           & \multicolumn{1}{c|}{30.79 / 0.8985} & \multicolumn{1}{c|}{29.67 / 0.8922} & 29.86 / 0.7895 \\
BSD\_All                              & \multicolumn{1}{c|}{28.66 / 0.8589} & \multicolumn{1}{c|}{33.35 / 0.9348} & 30.89 / 0.8049 \\
RSBlur                                & \multicolumn{1}{c|}{29.86 / 0.8855} & \multicolumn{1}{c|}{30.85 / 0.9069} & 32.53 / 0.8398 \\ \hline
\end{tabular}
\end{table*}
\setlength{\tabcolsep}{1.4pt}

\section{Limited Coverage of Real Datasets}

In the main paper, we show the limited coverage of the existing real datasets including RealBlur~\cite{jsrim-ECCV2020} and BSD\_All~\cite{Zhong_2020_ECCV,Zhong_2021_arxiv}.
Specifically, we train SRN-DeblurNet~\cite{Tao-CVPR18} using one dataset, and evaluate its performance on the other dataset.
In this supplementary material, we report a full comparison result among the real datasets including the RSBlur real dataset in \Tbl{Limited_coverage_realdataset}.
As the table shows, the deblurring performance of one dataset significantly drops on the other datasets.
This result again verifies the limited coverage of the existing real datasets and the usefulness of our synthesis pipeline.

\section{Saturation Synthesis: Local vs.~Global}

\begin{figure}[t]
\centering
\includegraphics[width=1\linewidth]{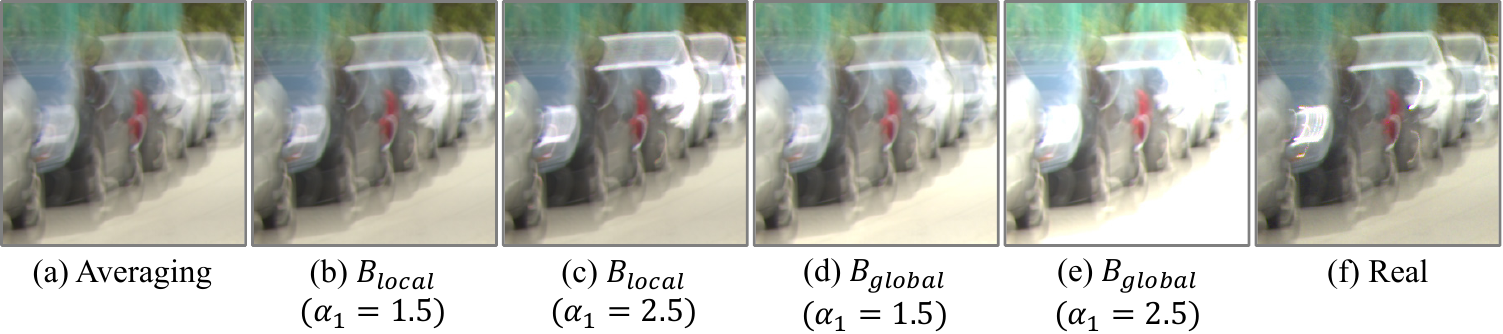}
\caption{Generated images from saturation synthesis methods using global scaling and local scaling.}
\label{fig:generating_saturation_synthesis}
\end{figure}

One important component in our blur synthesis pipeline is the saturation synthesis step, which locally scales intensity values of a blurred image before clipping as done in \cite{Hu-CVPR14}.
Another option that has been used in several previous works is global scaling~\cite{PAN_2017_TPAMI, Chen_2021_CVPR, NEURIPS2018_0aa1883c}.
In this section, we discuss the global and local scaling approaches and compare their performance.

Without considering noise and camera ISP for the brevity of the discussion, we can model a blurred image $B$ with clipped intensity values as $B=\textrm{clip}[I*K]$, where $\textrm{clip}[\cdot]$ is a clipping function, $K$ is a convolution kernel, and $I$ is a sharp image.
To obtain clipped intensity values in $B$, $I$ should contain unclipped intensity values larger than the upper limit of the dynamic range.
However, as sharp images also have the limited dynamic range, it is inevitable to synthesize sharp images with large intensity values, which can be done by either global scaling~\cite{PAN_2017_TPAMI, Chen_2021_CVPR, NEURIPS2018_0aa1883c} or local scaling~\cite{Hu-CVPR14}.

The global scaling approach generates saturated pixels by scaling all the intensity values as $\textrm{clip}[\alpha_1 \cdot I * K]$, where $\alpha_1$ is a scaling factor. The local scaling approach, on the other hand, scales only some intensity values as $\textrm{clip}[(I^{n\_sat} + \alpha_1 \cdot I^{sat}) * K]$, where $I^{n\_sat}$ and $I^{sat}$ are images that have non-zero pixels on non-saturated and saturated region of $I$, respectively.
Due to the information loss at clipped pixels, both methods randomly sample $\alpha_1$ to generate non-clipped pixels.
By replacing the convolution operation with $K$ by averaging operation over consecutive video frames, we can also model the blur synthesis based on averaging video frames, on which our analysis in the main paper is based.

\Fig{generating_saturation_synthesis} shows real saturated images and synthesized images using global scaling and local scaling.
Both methods cannot exactly reproduce the real saturated pixels (\Fig{generating_saturation_synthesis}(c),(e), and (f)) due to the missing information in sharp images caused by clipping. 
Nevertheless, the local scaling approach has a couple of advantages over the global scaling approach.
First, global scaling affects all the pixels, thus introduces a larger domain gap (brighter images) as shown in \Fig{generating_saturation_synthesis}(e), and severe distortion of the distribution of signal-dependent noise.
Second, as global scaling increases the intensities of larger areas, it is usually more difficult to mimic sharp light streaks often observed in real blurred images.
This difference can also affect the deblurring performance.
We conducted an experiment using the GoPro\_U training set with global scaling, and found that global scaling performs worse than our saturation synthesis by 0.35 dB on the RealBlur test set.

\section{Direct Measurement of the Quality of Blur Synthesis}

To evaluate the synthesis methods, we measure the deblurring performance trained with them in the main paper.
Another possible option would be to directly compare synthetic blurred images against real blurred images to quantify the quality of the synthetic methods using the RSBlur dataset.
In this section, we discuss about the direct measurement of the quality of blur synthesis and report additional evaluation results.

To measure the quality of different blur synthesis approaches, we measure the PSNR and SSIM values of synthetic blurred images against real-blurred images.
However, the PSNR or SSIM values of synthetic images are not 100\% reliable due to noise.
One possible option is to measure the KL-divergence of the distributions of real and synthesized images as done in noise-synthesis approaches~\cite{Abdelhamed_2019_ICCV, Chang_2020_ECCV, Jang_2021_ICCV}.
Specifically, we compute the KL-divergence between pixel-wise marginal distributions $(B_{real} - B_{interp})$ and $(B_{syn} - B_{interp})$ where $B_{real}$ and $B_{interp}$ is a real blurred image and an averaging of interpolated images, respectively. $B_{syn}$ is a synthesized blurred image using our pipeline.
We use $B_{interp}$ as rough noise-free counterparts of real blurred images for computing the KL-divergence.

\Tbl{additional_metrics} shows PSNR, SSIM, and the KL-divergence of synthesized images using variants of our synthesis methods. 
The table shows that frame interpolation and saturation synthesis are effective in terms of PSNR (methods 2 and 3). The method 7 shows noise and saturation synthesis methods improve the KL-divergence. We omit the KD-divergence values of methods 2 and 3 as the synthetic images have no noise thus their noise distributions are not properly defined.
Note that, measuring the KL-divergence also has its own flaws as we don't have accurate noise-free counterparts of real blurred images. So, 100\% accurate estimation of real distribution is not feasible.
Also, as our ultimate goal is to improve the deblurring quality, the deblurring performance of the trained network is a most reasonable measure. 

\setlength{\tabcolsep}{7pt}
\begin{table}[t]
\centering
\caption{Comparison among different blur synthesis methods. We compute PSNR, SSIM and KL-divergence (KLD) from synthesized and real blurred images.}
\label{tbl:additional_metrics}
\scalebox{1.0}{
\begin{tabular}{|c|cccc|c|}
\hline
\multicolumn{5}{|c|}{Blur Synthesis Methods}                                                     &                           \\ \hline
No.                     & Interp.                   & Sat.   & Noise & ISP                       & PSNR / SSIM / KLD         \\ \hline
1                       &                           &        &       &                           & 35.74 / 0.8802 / 1.0064  \\
2                       & \checkmark &        &       &                           & 35.97 / 0.8888 / \ \ \ \ - \ \ \ \\
3                       & \checkmark & Ours &       &                           & 36.03 / 0.8888 / \ \ \ \ - \ \ \ \\
4                       & \checkmark &        & G     &                           & 34.16 / 0.8075 / 0.4312   \\
5                       & \checkmark &        & G+P   & \checkmark & 33.14 / 0.7928 / 0.3319   \\
6                       & \checkmark & Ours   & G     &                           & 34.21 / 0.8077 / 0.4313   \\
7                       & \checkmark & Ours   & G+P   & \checkmark & 33.18 / 0.7928 / 0.3263   \\ \hline
\end{tabular}
}
\end{table}

\section{Training with Both Real and Synthetic Datasets}

Even if a real-world blur training set is available, an additional synthetic dataset generated by our method could further improve the deblurring performance on real blurred images.
One important question when training with two datasets is how to mix them.
To verify the effect of using both real and synthetic datasets and the mixing strategy, in this section, we compare the deblurring performance obtained using one of real and synthetic datasets and using both of them.
Specifically, we compare four different training strategies: training 1) with only RealBlur\_J, 2) with only GoPro\_U, 3) with both RealBlur\_J and GoPro\_U together half and half at each iteration, and 4) with RealBlur\_J and GoPro\_U one by one at each iteration.
We train SRN-DeblurNet using the four strategies and evaluate their performances on the RealBlur test set.
As the table shows, using both RealBlur\_J and GoPro\_U clearly improves the deblurring performance regardless of the training strategies.

\setlength{\tabcolsep}{5pt}
\begin{table}[t]
\centering
\caption{Comparison of different training strategies using additional synthetic datasets for further performance improvements.}
\label{tbl:training_process}
\scalebox{1.0}{
\begin{tabular}{|c|cc|c|}
\hline
No. & Train dataset & Strategy      & PSNR / SSIM    \\ \hline
1   & RealBlur\_J          &             & 30.79 / 0.8985 \\
2   & GoPro\_U           &             & 30.75 / 0.9019 \\
3   & RealBlur\_J + GoPro\_U    &       Half \& Half      & 31.17 / 0.9065 \\
4   & RealBlur\_J + GoPro\_U    & One $\times$ One & 31.15 / 0.9059 \\ \hline
\end{tabular}
}
\end{table}

\section{Additional Results on Other Datasets}

In the main manuscript, we evaluate the proposed pipeline on the RealBlur~\cite{jsrim-ECCV2020} and BSD\_All~\cite{Zhong_2020_ECCV,Zhong_2021_arxiv} datasets.
Additionally, in this supplementary material, we also report evaluation results on K\"{o}hler~\etal's dataset~\cite{Kohler-ECCV12}.
\Tbl{kohler_result} shows the performance of SRN-DeblurNet~\cite{Tao-CVPR18} trained with variants of the proposed convolution-based synthesis pipeline.
Note that the images in K\"{o}hler~\etal's dataset are in the linear RGB space, and do not have saturated pixels.
Thus, the method 5 performs the best, while the other methods with wrong CRFs show significant performance drops.
Again, the poor performances of the existing real datasets on K\"{o}hler~\etal's dataset prove their limited coverage, as discussed in the main manuscript.

\setlength{\tabcolsep}{4pt}
\begin{table*}[ht]
\centering
\caption{Performance comparison of different blur synthesis methods on the K\"{o}hler \etal's~\cite{Kohler-ECCV12} dataset. Sat.: Saturation synthesis. sRGB: Gamma correction of sRGB space. G: Gaussian noise. G+P: Gaussian and Poisson noise. A7R3: Using camera ISP parameters estimated from a Sony A7R3 camera, which was used for collecting the RealBlur dataset.}
\label{tbl:kohler_result}
\begin{tabular}{|cccccc|c|}
\hline
\multicolumn{6}{|c|}{Blur Synthesis Methods}     & PSNR / MSSIM   \\ \hline
No. & Training set & CRF    & Sat. & Noise & ISP & K\"{o}hler \etal             \\ \hline
1   & RealBlur\_J  &        &      &       &     & 26.79 / 0.8401 \\
2   & BSD\_All     &        &      &       &     & 25.24 / 0.7920 \\
3   & RSBlur       &        &      &       &     & 26.29 / 0.8285 \\
4   & GoPro\_U     & Linear &      &       &     & 28.28 / 0.8599 \\
5   & GoPro\_U     & Linear &      & G     &     & 28.41 / 0.8624 \\
6   & GoPro\_U     & sRGB    &      &       &     & 26.86 / 0.8430 \\
7   & GoPro\_U     & sRGB    &      & G     &     & 27.23 / 0.8529 \\
8   & GoPro\_U     & sRGB    & Ours & G     &     & 27.10 / 0.8500 \\
9   & GoPro\_U     & sRGB, A7R3 & Ours & G+P   &A7R3 & 27.41 / 0.8516 \\ \hline
\end{tabular}
\end{table*}

\begin{figure}[t]
\centering
\includegraphics[width=1\linewidth]{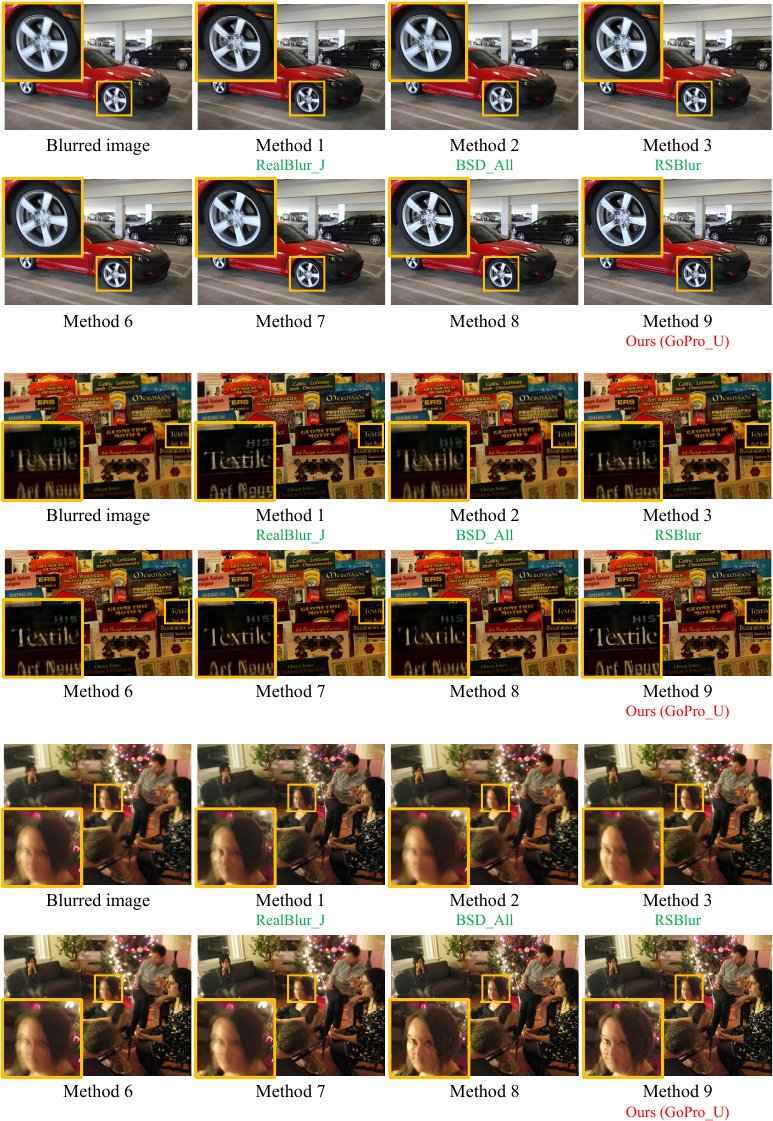}
\vspace{-0.8cm}
\caption{Qualitative comparison of deblurring results on the Lai~\etal's~dataset~\cite{Lai-CVPR16} produced by models trained with different synthesis methods in \Tbl{kohler_result}.
Best viewed in zoom in.
}
\label{fig:qualitative_lai}
\end{figure}

\begin{figure}[t]
\centering
\includegraphics[width=1\linewidth]{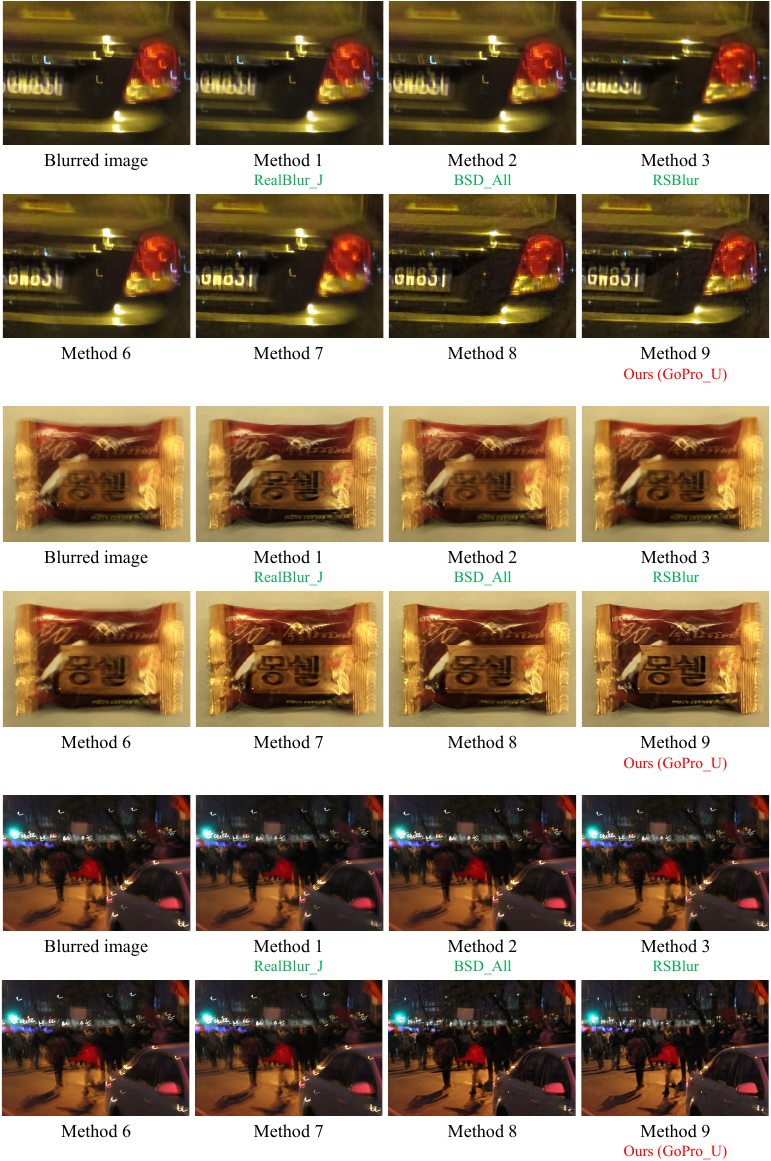}
\vspace{-0.8cm}
\caption{Qualitative comparison of deblurring results on the Lai~\etal's~dataset~\cite{Lai-CVPR16} produced by models trained with different synthesis methods in \Tbl{kohler_result}.
Best viewed in zoom in.
}
\label{fig:qualitative_lai2}
\end{figure}

We also compare variants of the convolution-based synthesis pipeline in \Tbl{kohler_result} on Lai~\etal's dataset~\cite{Lai-CVPR16}, which provides 100 real blurred images without ground-truth sharp images for qualitative comparison. 
In \Figs{qualitative_lai} and \ref{fig:qualitative_lai2}, methods 1, 2, and 3 show that the models trained on real datasets do not successfully deblur real blurred images of Lai~\etal's dataset. 
On the other hand, the method 9 shows that the deblurring results produced by the model trained with our pipeline performs better.
This again shows the limited coverage of the real datasets and the practicality of the proposed pipeline.
The proposed pipeline can generate a huge number of images with various contents, and blur shapes and sizes effortlessly compared to collecting real datasets, leading to better deblurring performance.

\section{Additional Qualitative Examples}

\begin{figure*}[!t]
\centering
\includegraphics[width=0.95\linewidth]{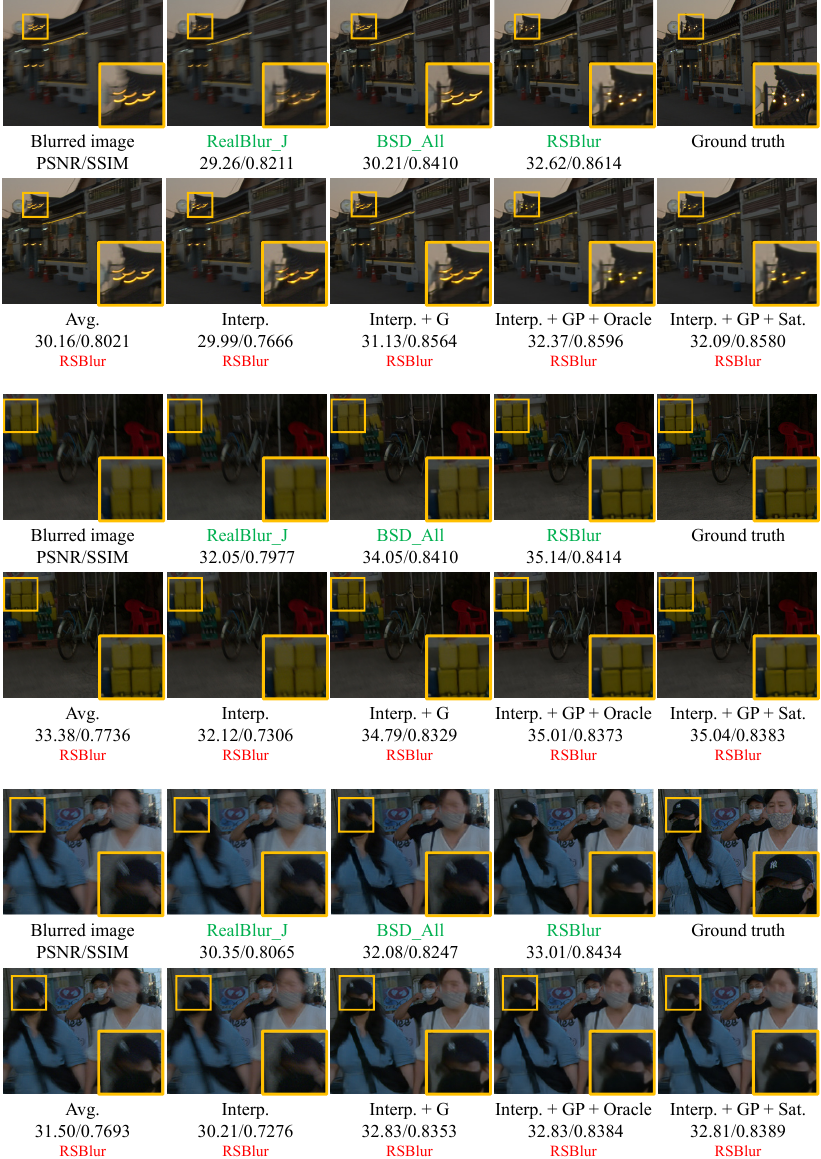}
\caption{Qualitative comparison on the RSBlur dataset.
\textcolor{green}{Green}: Trained on real blurred images.
\textcolor{red}{Red}: Trained on synthetic blurred images.
Avg.: Na\"{i}ve averaging-based blur synthesis.
Interp.: Averaging-based blur synthesis using frame interpolation.  
G: Gaussian noise. 
Oracle: Using oracle saturated images.
Sat: Our saturation synthesis. 
All of synthesis methods consider gamma decoding and encoding.}
\label{fig:RSBlur_results}
\end{figure*}

\begin{figure*}[!t]
\centering
\includegraphics[width=0.95\linewidth]{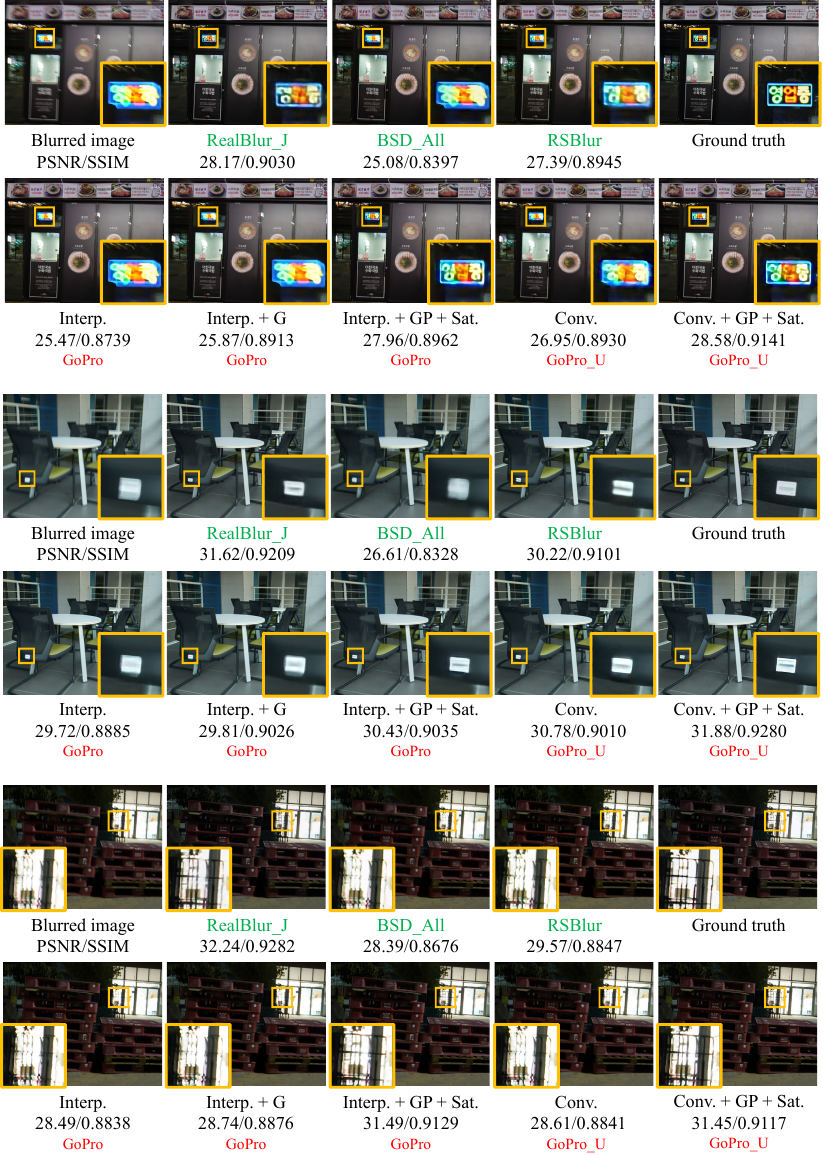}
\caption{Qualitative comparison on the RealBlur\_J dataset~\cite{jsrim-ECCV2020}.
\textcolor{green}{Green}: Trained on real blurred images.
\textcolor{red}{Red}: Trained on synthetic blurred images.
Interp.: Averaging-based blur synthesis using frame interpolation. 
G: Gaussian noise. 
GP: Gaussian and Poisson noise with a camera ISP of Sony A7R3. 
Sat: Our saturation synthesis. 
Conv.: Convolution-based blur synthesis.
All of synthesis methods consider gamma decoding and encoding.}
\label{fig:RealBlur_results}
\end{figure*}

\begin{figure*}[!t]
\centering
\includegraphics[width=0.95\linewidth]{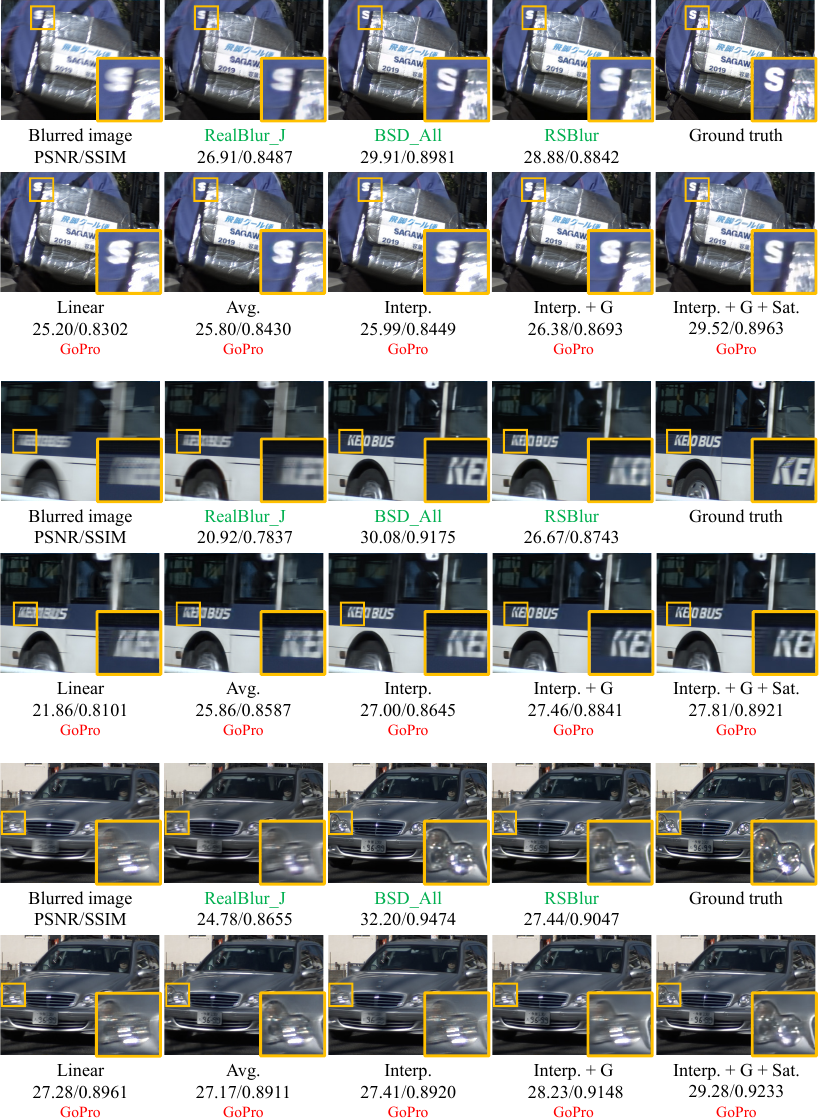}
\caption{Qualitative comparison on the BSD\_All dataset~\cite{Zhong_2020_ECCV,Zhong_2021_arxiv}.
\textcolor{green}{Green}: Trained on real blurred images.
\textcolor{red}{Red}: Trained on synthetic blurred images.
Linear: Na\"{i}ve averaging-based blur synthesis with linear CRF.
Avg.: Na\"{i}ve averaging-based blur synthesis.
Interp.: Averaging-based blur synthesis using frame interpolation.  
G: Gaussian noise. 
Sat: Our saturation synthesis. 
All of synthesis methods except Linear consider gamma decoding and encoding.}
\label{fig:BSD_results}
\end{figure*}

\Figs{RSBlur_results}, \ref{fig:RealBlur_results} and \ref{fig:BSD_results} show additional qualitative examples on the RSBlur, RealBlur\_J~\cite{jsrim-ECCV2020}, and BSD\_All~\cite{Zhong_2020_ECCV,Zhong_2021_arxiv} datasets, respectively.
The figures show deblurring results of SRN-DeblurNet~\cite{Tao-CVPR18} trained with training images synthesized by different methods in order to compare different blur synthesis methods.

